\documentclass[10pt,journal,compsoc]{IEEEtran}
\usepackage{stfloats}
\usepackage{color}
\usepackage{float}
\usepackage{lineno,hyperref,amsmath}
\usepackage{graphicx,float}
\usepackage{subfig}
\usepackage{breqn}
\DeclareMathOperator*{\argmax}{argmax}
\ifCLASSOPTIONcompsoc
  \usepackage[nocompress]{cite}
\else
  \usepackage{cite}
\fi
\begin{document}
%
\title{An End-to-End Deep Learning Histochemical Scoring System for Breast Cancer Tissue Microarray}
%
%
%
%
%

\author{Jingxin Liu, Bolei Xu, Chi Zheng, Yuanhao Gong, Jon Garibaldi, Daniele Soria, Andew Green, Ian O. Ellis, Wenbin Zou, Guoping Qiu

\IEEEcompsocitemizethanks{\IEEEcompsocthanksitem J. Liu, B. Xu, W. Zou, and G. Qiu are with the College of Information Engineering, Shenzhen University, China; G. Qiu is also with School of Computer Science, The University of Nottingham, UK
\IEEEcompsocthanksitem C. Zheng is with Ningbo Yongxin Optics Co., LTD, Zhejiang, China
\IEEEcompsocthanksitem Y. Gong is with Computer Vision Laboratory , ETH Zurich, Switzerland
\IEEEcompsocthanksitem J. Garibaldi is with School of Computer Science, The University of Nottingham, UK
\IEEEcompsocthanksitem D. Soria is with Department of Computer Science, The University of Westerminster, UK
\IEEEcompsocthanksitem A. Green and I. O. Ellis are with Faculty of Medicine \& Health Sciences, The University of Nottingham, United Kingdom.

\IEEEcompsocthanksitem G. Qiu is the corresponding author, E-mail:qiu@szu.edu.cn}
}

%
%

\markboth{Journal of \LaTeX\ Class Files,~Vol.~14, No.~8, August~2015}%
{Shell \MakeLowercase{\textit{et al.}}: Bare Demo of IEEEtran.cls for Computer Society Journals}
%



\IEEEtitleabstractindextext{%
\begin{abstract}
One of the methods for stratifying different molecular classes of breast cancer is the Nottingham Prognostic Index Plus (NPI+) which uses breast cancer relevant biomarkers to stain tumour tissues prepared on tissue microarray (TMA). To determine the molecular class of the tumour, pathologists will have to manually mark the nuclei activity biomarkers through a microscope and use a semi-quantitative assessment method to assign a histochemical score (H-Score) to each TMA core. However, manually marking positively stained nuclei is a time consuming, imprecise and subjective process which will lead to inter-observer and intra-observer discrepancies. In this paper, we present an end-to-end deep learning system which directly predicts the H-Score automatically. The innovative characteristics of our method is that it is inspired by the H-Scoring process of the pathologists where they count the total number of cells, the number of tumour cells, and categorise the cells based on the intensity of their positive stains. Our system  imitates the pathologists' decision process and uses one fully convolutional network (FCN) to extract all nuclei region (tumour and non-tumour), a second FCN to extract tumour nuclei region, and a multi-column convolutional neural network which takes the outputs of the first two FCNs and the stain intensity description image as input and acts as the high-level decision making mechanism to directly output the H-Score of the input TMA image. In additional to developing the deep learning framework, we also present methods for constructing positive stain intensity description image and for handling discrete scores with numerical gaps. Whilst deep learning has been widely applied in digital pathology image analysis, to the best of our knowledge, this is the first end-to-end system that takes a TMA image as input and directly outputs a clinical score. We will present experimental results which demonstrate that the H-Scores predicted by our model have very high and statistically significant correlation with experienced pathologists' scores and that the H-Scoring discrepancy between our algorithm and the pathologits is on par with that between the pathologists. Although it is still a long way from clinical use, this work demonstrates the possibility of using deep learning techniques to automatically and directly predicting the clinical scores of digital pathology images. 
\end{abstract}

\begin{IEEEkeywords}
H-Score, Immunohistochemistry, Diaminobenzidine, Convolutional Neural Network, Breast Cancer
\end{IEEEkeywords}}

\maketitle

\IEEEdisplaynontitleabstractindextext

%
\IEEEpeerreviewmaketitle

\IEEEraisesectionheading{\section{Introduction}\label{sec:introduction}}
\label{sec:intro}
Breast cancer (BC) is a heterogeneous group of tumours with varied genotype and phenotype features \cite{rakha2014nottingham}. Recent research of Gene Expression Profiling (GEP) suggests that BC can be divided into distinct molecular tumour groups \cite{perou1999distinctive, nielsen2010comparison}. Personalised BC management often utilizes robust commonplace technology such as immunohistochemistry (IHC) for tumour molecular profiling  \cite{soria2010methodology,green2013identification}.

\begin{figure*}[t]
	\centering
	\includegraphics[width=\textwidth]{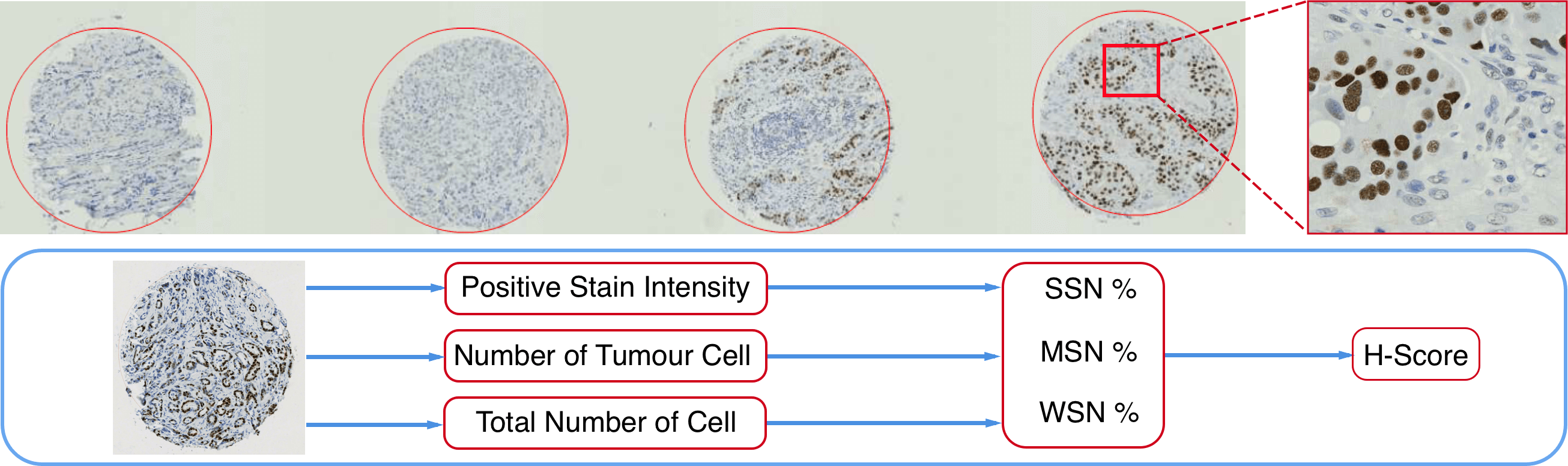}
	\caption{Top: Example images extracted from digital TMA slides. Each red circle contains one TMA core stained by Diaminobenzidine-Hematoxylin (DAB-H). The brown colours indicate positive and the blue colours indicate negative. Bottom: A schematic illustration of the traditional manual H-Scoring procedure: It needs to first count the total number of nuclei, then the number of strongly stained, moderately stained and weakly stained tumour nuclei, respectively. The final H-Score is then calculated according Eq.\ref{eq:hscore}.}
	\label{fig:teaser}
\end{figure*}

Diaminobenzidine (DAB) based IHC techniques stain the target antigens (detected by biomarkers) with brown colouration (positive) against a blue colouration (negative) counter-stained by Hematoxylin (see Fig.\ref{fig:teaser} for some example images). To determine the biological class of the tumour, pathologists will mark the nuclei activity biomarkers through a microscope and give a score based on a semi-quantitative assessment method called the modified histochemical scoring (H-Score) \cite{mccarty1985estrogen, goulding1995new}. The H-Scores of tissue samples stained with different biomarkers are combined together to determine the biological class of a case. Clinical decision making is to choose an appropriate treatment from a number of available treatment options according to the biological class of the tumour. For instance, one of the methods for stratifying different molecular classes is the Nottingham Prognosis Index Plus (NPI +)\cite{rakha2014nottingham} which uses 10 breast cancer relevant biomarkers to stain tumour tissues prepared on tissue microarray (TMA). Tissue samples stained by each of these 10 biomarkers are given a histochemical score (H-Score) and these 10 scores together will determine the biological class of the case. 

Therefore, H-Score is one of the most important pieces of information for molecular tumour classification. When the tumour region occupies more than 15\% of the TMA section, a H-Score is calculated based on a linear combination of the percentage of strongly stained nuclei ($SSN$), the percentage of moderately stained nuclei ($MSN$) and the percentage of weakly stained nuclei ($WSN$) according to equation (1):
\begin{equation}
\label{eq:hscore}
\rm{H-Score} = 1 \times WSN + 2 \times MSN + 3 \times SSN 
\end{equation}
The final score has a numerical value ranges from 0 to 300. Thus, the histochemical assessment of the TMA's is based on the following semi-quantitative information: the total number of cells, the number of tumour cells and the stain intensity distributions within the tumour cells. 
In clinical practice, diagnosis requires averaging two experienced pathologists' assessments. 
Manually marking the positively stained nuclei is obviously a time consuming process. As visual assessment of the TMA's is subjective, there is the problem of inter-observer discrepancy and the issue of repeatability. The semi-quantitative nature of the method (strongly stained, moderately stained and weakly stained, the definitions of strong, moderate and weak cannot be precise and subjective), makes it even more difficult to ensure inter-subject as well as intra-subject consistency.   

With the increasing application of clinicopathologic prognosis, Computer Aided Diagnosis (CAD) systems have been proposed to support the pathologists' decision making. 
The key parameters in tissue image assessment include the number of tumour cells, the positive staining intensities within these cells and the total number of all cells in the image. To classify the positively stained pixels and their stain intensity, methods such as colour deconvolution that perform mathematical transformation of the RGB image \cite{ruifrok1997quantification} \cite{ruifrok2001quantification} are widely used to separate positive stains from negative stains. Numerous computer-assisted approaches have been proposed for cell or nuclei detection and segmentation \cite{irshad2014methods}. Most literature on histopathology image analysis perform various low-level quantification steps, there is still little attempt to perform end-to-end assessment of the image directly. 

In this paper, we ask this question: is it possible to develop a CAD model that would directly give a high-level assessment of a digital pathological image, just like an experienced pathologist would, for example, to give out a H-Score directly? 
In an attempt to answer this question, we propose an end-to-end deep learning system for directly predicting the H-Scores of breast cancer TMA images, see Fig. \ref{fig:overall_flow}. Instead of pushing the raw digital images into the neural network directly, we follow a similar process that pathologists use for H-Score estimation. We first construct a stain intensity nuclei image (SINI) which only contains nuclei pixels and their corresponding stain intensity information, and a stain intensity tumour image (SITI) which only contains tumour nuclei pixels and their corresponding stain intensity information. The SINI and SITI block irrelevant background pixels while only retain useful information for calculating the H-Score. 
These two H-Score relevant images are then fed into a dual-channel convolutional neural network with two input pipelines, which are finally merged into one pipeline to give an output (H-Score). 
To the best of our knowledge, this is a first work that attempts to develop deep learning based TMA processing model that directly outputs the histochemical scores. We will present experimental results which demonstrate that the H-Scores predicted by our model have high and statistically significant correlation with experienced pathologists' scores and that the H-Scoring discrepancy between our algorithm and the pathologists is on par with that between the pathologists. Although it is still perhaps a long way from clinical use, this work nevertheless demonstrates the possibility of automatically scoring cancer TMA's based on deep learning. 

\section{Related Works}
Researchers have proposed various computer-assisted analysis methods for histopathological images \cite{kothari2013pathology}. For pixel-level positive stain segmentation, Pham \cite{pham2007quantitative} adapted Yellow channel in CMYK model, which is believed to have strong correlation with the DAB stain; Ruifrok \cite{ruifrok1997quantification} presented the brown image calculated based on mathematical transformation of the RGB image. Yao \cite{yao2012interactive} employed Hough forest for mitotic cell detection, which is a combination of generalized Hough transform and random decision trees. Shu \emph{et al.} \cite{shu2013segmenting} proposed utilizing morphological filtering and seeded watershed for overlapping nuclei segmentation. Object-based CAD systems have also been developed for tubule detection in breast cancer \cite{basavanhally2011incorporating}, glandular structure segmentation \cite{fu2014novel}, and etc.

With the development of deep learning techniques, various deep neural network based CAD models have been published. Deep convolutional networks with deeper architectures can be used to build more complex models which will result in more powerful solutions. Li \cite{li2017hep} used a 88-layer residual network for human epithelial type 2 (HEp-2) cell segmentation and classification. \emph{AggNet} with a novel aggregation layer is proposed for mitosis detection in breast cancer histology images \cite{albarqouni2016aggnet}. Google brain presented a multi scale CNN model to aid breast cancer metastasis detection in lymph nodes\cite{liu2017detecting}. A deep learning-based system is proposed for the detection of metastatic cancer from whole slide images, which won the Camelyon Grand Challenge 2016 \cite{wang2016deep}. Shah \emph{et al.} \cite{shah2016deep} presented the first completely data-driven model integrated numerous biologically salient classifiers for invasive breast cancer prognosis. A symmetric fully convolutional network is proposed by Ronneberger for microscopy image segmentation \cite{ronneberger2015u}. 

Digital pathology is relative new compared with other type of medical imaging such as X-ray, MRI, and CT. Deep learning as one of the most powerful machine learning techniques emerged in recent years has seen widespread applications in many areas. Yap \emph{et al.} \cite{yap2017automated} investigated  three deep learning models for breast ultrasound lesion detection. Moeskops \cite{moeskops2016deep} introduced a single CNN model with triplanar input patches for segmenting three different types of medical images, brain MRI, breast MRI and cardiac CTA. A combination of multi-channel image representation and unsupervised candidate proposals is proposed for automatic lesion detection in breast MRI \cite{amit2017hybrid}. 

Most existing high-level CAD frameworks directly follow the assessment criteria by extracting quantitative information from the digital images. Masmoudi \emph{et al.}\cite{masmoudi2009automated} proposed an automatic Human Epidermal Growth Factor Receptor 2 (HER2) assessment method, which is an assemble algorithm of colour pixel classification, nuclei segmentation and cell membrane modelling. Gaussian-based bar filter was used for membrane isolation after colour decomposition in \cite{hall2008computer}. Trahearn \emph{et al.} \cite{trahearn2016simultaneous} established a two-stage registration process for IHC stained WSI scoring. Thresholds were defined for DAB stain intensity groups, and tumour region and nuclei were detected by two different detectors. Recently, Zhu \cite{zhu2017sur} proposed to train an aggregation model based on deep convolutional network for patient survival status prediction.  

\section{Problem and Method}
An immunohistochemical assessment can be formulated as a model $\mathsf{F}$ that maps the input images from the input space $\mathcal{I}$ to the a label space $\mathcal{L}$. Given an input image $I\in \mathcal{I}$, its label $l \in \mathcal{L}$ is assigned according to the quantitative information of positive staining intensity $P_s$, the number of tumour cells $N_{t}$ and total number of cells $N_{e}$ in the image $x$:
\begin{equation}
l = \mathsf{F}(I|P_s, N_t, N_e),
\end{equation}

\begin{figure}[t]
	\centering
	\begin{tabular}{ccc}
		\includegraphics[width=2.5cm,height = 2.5cm]{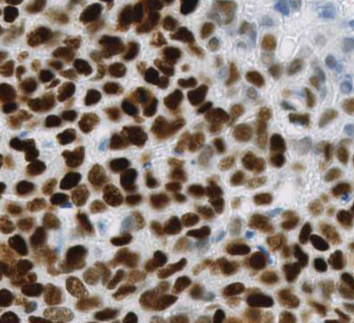}&
		\includegraphics[width=2.5cm,height = 2.5cm]{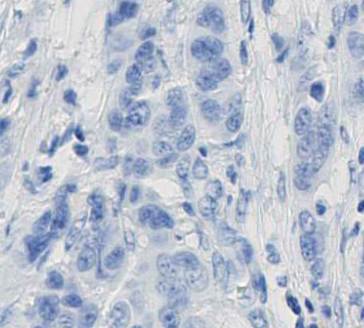}&
		\includegraphics[width=2.5cm,height = 2.5cm]{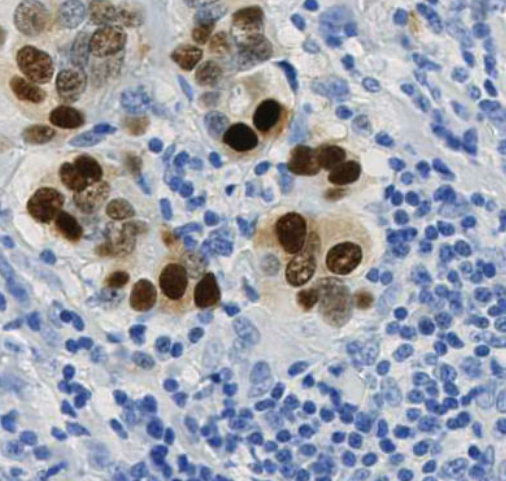}\\
		(a)&(b)&(c)
	\end{tabular}
	\caption{Examples of challenging cases of quantitative measurement of biomarkers based on visual assessment. (a) A variety of stain intensities; (b) unclear staining and overlapping of nucleus; (c) Size differences between different type of nucleus.}
	\label{fig:problem}
\end{figure}

Traditional assessment methods have at least three unsolved issues for both the pathologists and the CAD systems. Firstly, the positive staining intensity needs to be categorized into four classes: \emph{unstained}, \emph{weak}, \emph{moderate}, and \emph{strong}. However, there is no standard quantitative criterion for classifying the DAB stain intensity. Thus, two pathologists often classify the same staining intensity into two different categories or two different intensities into the same category. Furthermore, the human visual system may pay more attention to strongly stained regions but they are often surrounded by a variety of staining intensities \cite{trahearn2016simultaneous}, which may also affect the assessment results. Secondly, cell/nuclei instance counting is a very important parameter in the assessment. Nevertheless, both human and computer still cannot deal with the difficulty of counting overlapping cells very well. Moreover, variability in the appearance of different types of nucleus, heterogeneous staining, and the complex tissue architectures make individually segmenting cell/nuclei a very challenging problem. Thirdly, the apparent size differences between tumour nuclei and normal nuclei will affect the quantitative judgement of tumour nuclei assessment. Examples of these challenging cases are illustrated in Fig. \ref{fig:problem}.

\begin{figure*}[t]
	\centering
	\includegraphics[width=17.5cm,height = 5.5cm]{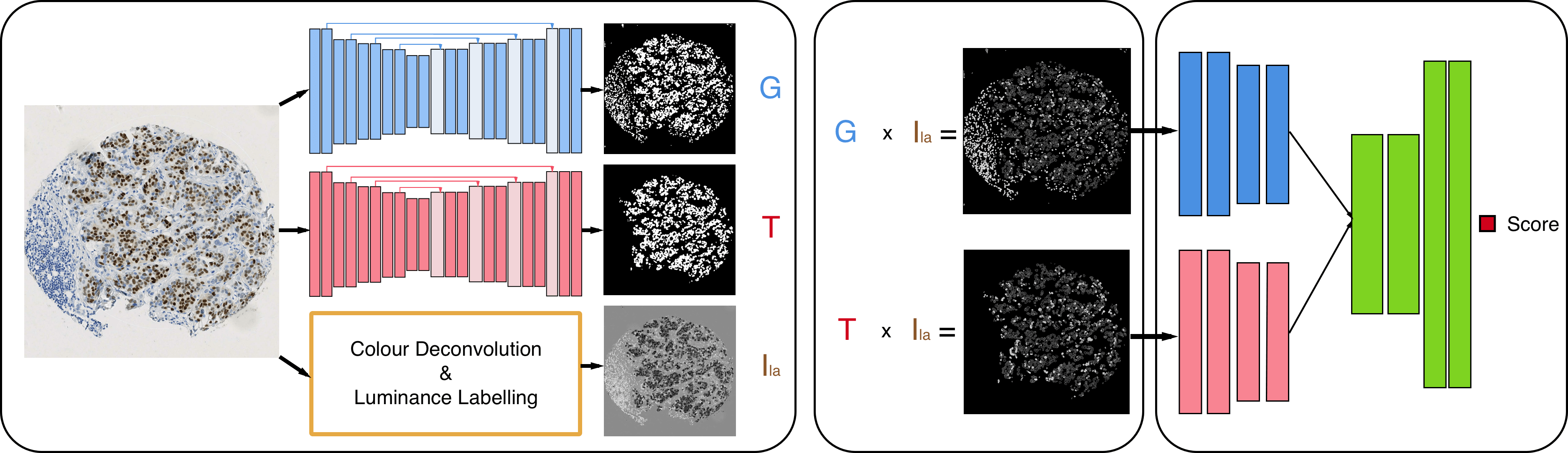}
	\caption{The overview of our proposed H-Score prediction framework. An input TMA image is first processed by two FCNs to extract tumour cells and all cells (tumour and non-tumour) to produce two mask images. The input image is also processed by colour deconvolution and positive stain classification to output a stain intensity description image. The two mask images are used to filter out irrelevant information in the stain intensity description image and only the useful information is fed to a deep convolutional neural network for the prediction of H-Score of the input TMA.}
	\label{fig:overall_flow}
\end{figure*}

To tackle the problem mentioned above, we propose to develop a convolutional neural network (CNN) based CAD framework for biomarker assessment of TMA images. Instead of using CNN as a feature extractor or for low level processing such as cell segmentation only, we have developed an end-to-end system which directly predicts the biomarker score (H-Score). The innovative characteristic of our method is that it is inspired by the H-Scoring process of the pathologists where they count the total number of nuclei and the number of tumour nuclei and categorise tumour nucleus based on the intensity of their positive stains. In the complete system, as illustrated in Fig. \ref{fig:overall_flow}, one fully convolutional network (FCN) is used to extract all nuclei region which acts as the step of counting all nucleus and capture all foreground information, another FCN is used to extract tumour nuclei region which acts as the step of counting all tumour nucleus. To mimic the process of categorising tumour nuclei based on their positive stain intensities, we derive a stain intensity image which together with the outputs of the two FCNs are presented to another deep learning network which acts as the high-level decision making mechanism to directly output the H-Score of the input TMA image.  

\subsection{Stain Intensity Description}
\label{sec:sid}
\begin{figure*}[t]
	\centering
	\includegraphics[width=12.5cm,height = 4.5cm]{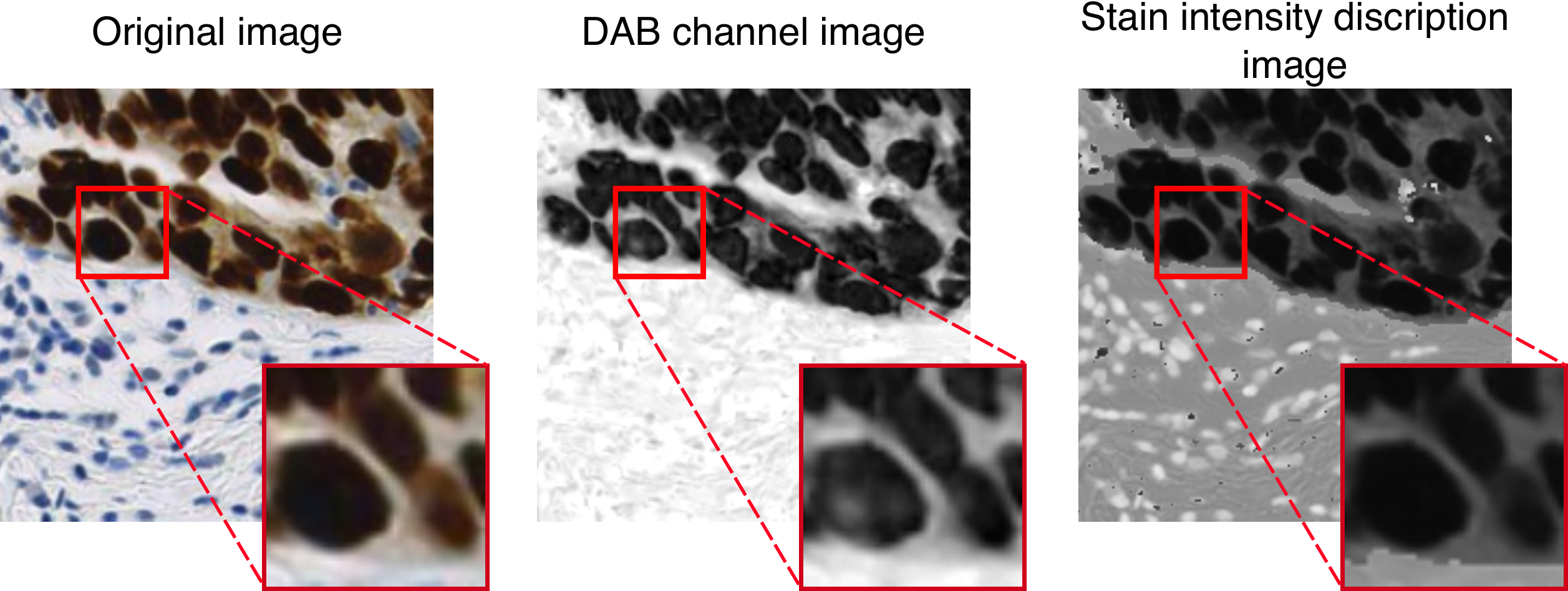}
	\caption{A comparison of different images generated during the process of 
		stain intensity description. The highlighted subimage contains strongly stained nuclei.} 
	\label{fig:sid}
\end{figure*}

Although various DAB stain separation methods have been proposed \cite{ruifrok2001quantification,brey2003automated}, few work studied the stain intensity description and grouping. Since there is no formal definitions for the boundaries between stain intensity groups (e.g, \emph{strong}, \emph{moderate}, \emph{weak}), previous works used manually defined thresholds for pixel-wise classification to segment positive stains into each stain group \cite{trahearn2016simultaneous}. 

\begin{figure}[!th]
	\centering
	\includegraphics[height = 5.0cm]{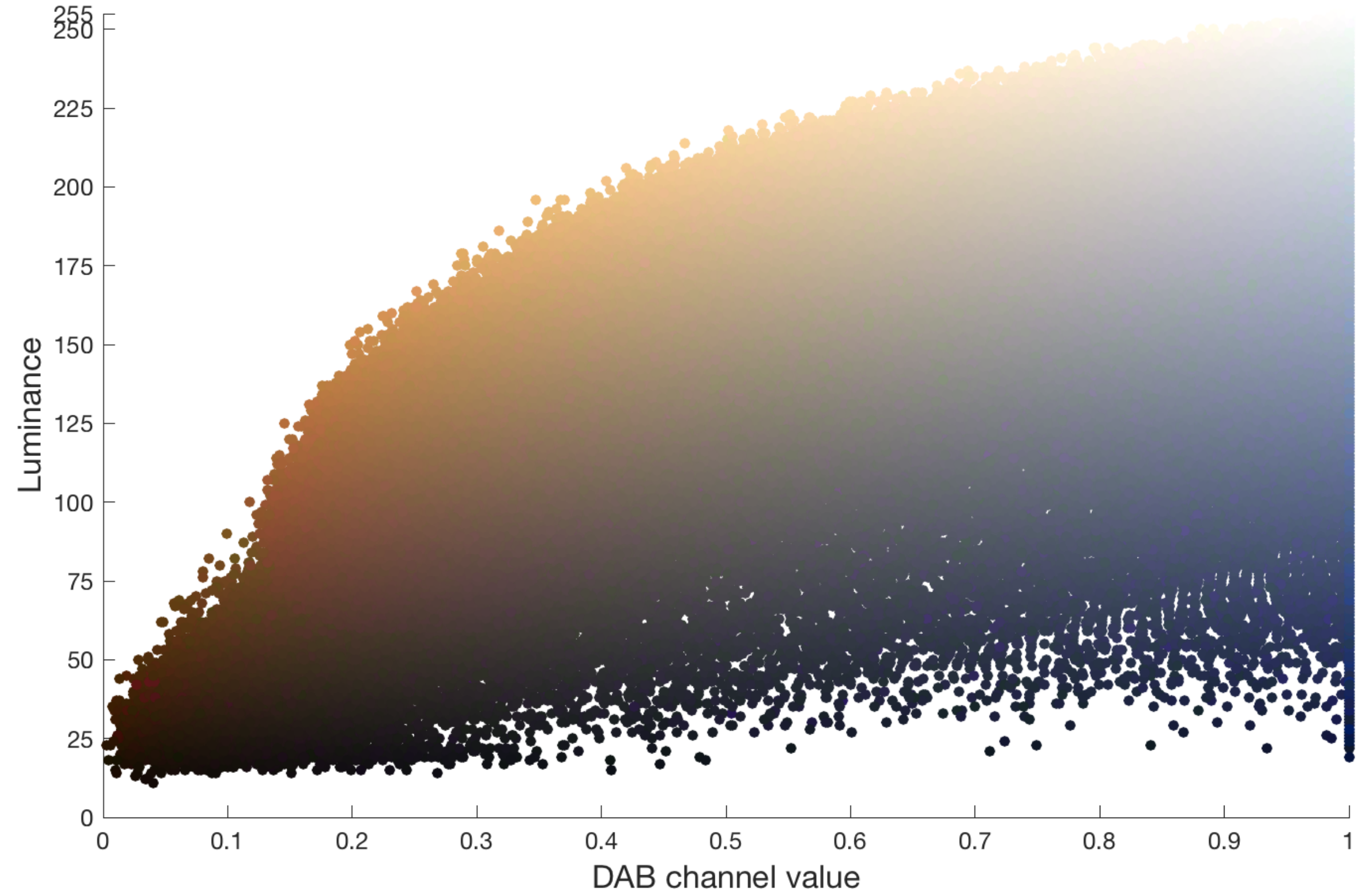}
	\caption{Visualization of pixel colours of the DAB-H images along the luminance axis and the colour deconvolution DAB axis.}
	\label{fig:cd}
\end{figure}

In this work, we propose to directly use the luminance values of the image to describe the staining intensity instead of setting artificial intensity category boundaries. The original RGB image $I$ is first transformed into three-channel stain component image ($I_{_{DAB-H}}=[I_{_{DAB}},I_{_{H}},I_{_{Other}}]$) using colour deconvolution \cite{ruifrok2001quantification}:
\begin{equation}
I_{_{DAB-H}}= M^{-1} I_{OD},
\end{equation}
where $M$ is the stain matrix composed of staining colours equal to
\begin{equation}
\left[
\begin{matrix}
0.268 &0.570&0.776 \\
0.650&0.704&0.286 \\
0.0&0.0&0.0
\end{matrix}
\right]
\end{equation}
for DAB-H stained images, and $I_{OD}$ is Optical Density converted image calculated according Lambert-Beers law:
\begin{equation}
I_{OD}= -log(\frac{I}{I_{0}}),
\end{equation}
$I_{0} = [255,255,255]$ is the spectral radiation intensity for a typical 8bit RGB camera \cite{haub2015model}. 

\begin{figure*}[!hb]
	\centering	
	\includegraphics[height = 1.3cm]{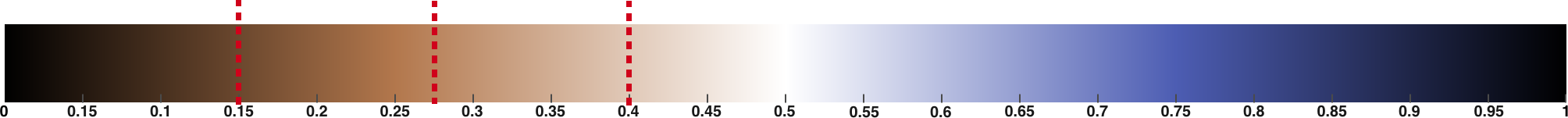}		
	\caption{An illustration of the value of $l_{la}$ and its corresponding stain intensity. The red dot lines are the thresholds of stain intensity groups \cite{trahearn2016simultaneous}.}
	\label{fig:color_thres}
\end{figure*}
Only the DAB channel image $I_{_{DAB}}$ from the three colour deconvolution output channels is used, which describes the DAB stain according the chroma difference. 

Most previous works set a single threshold on $I_{_{DAB}}$ to separate positively stained tissues. However, as shown in Fig.\ref{fig:sid}, the deeply stained positive nuclei can have dark and light pixel values on the DAB channel image, since the strongly stained pixels will have significantly broader hue spectrum. Furthermore, as illustrated in Fig.\ref{fig:cd}, the same DAB channel value can correspond to different pixel colours. Also, from Fig.\ref{fig:cd}, it is clear that in order to separate the positive stain (brown colour) from the negative stain (blue colour), the DAB channel thresholds should be set based on the luminance values. In this paper, we use the Luminance Adaptive Multi-Thresholding (LAMT) method developed by the authors \cite{liu2016luminance} to classify positively stained pixels. Specifically, the transformed pixel $I_{_{DAB}}(m,n)$ is divided into $K$ equal intervals according to the luminance:
\begin{equation}
\footnotesize
I_{_{DAB}}^{k}(m,n) = \{I_{_{DAB}}^{k}(m,n) \in I_{_{DAB}})| \xi^{k} < I_l(m,n) \leq \zeta^{k}\}
\end{equation}
where $k = 1,2,...,K$; $\xi_i$ and $\zeta_i$ are lower and upper boundary respectively of $k$th luminance interval. $I_l$ is the luminance image of the original RGB image calculated according to Rec. 601 \cite{rec1995bt}:
\begin{equation}
I_l  = 0.299 \times I_R+ 0.587 \times I_G + 0.114 \times I_B. 
\end{equation}
The transformed pixels are thresholded with different values according to its luminance instead of a single threshold, the threshold $t_k$ is assigned as follows:
\begin{equation}
t_k = \mathop{\argmax}_{c\in \mathcal{C}}{P(c|I_{_{DAB}}^{k}(m,n) )}
\end{equation}
where $\mathcal{C} = \{c_{_{DAB}},c_{_{H}}\}$ is the stain label.



Once we have separated the positive stain from the negative stain, we need to find a way to describe the stain intensity. As we have already seen in Fig.\ref{fig:sid} and Fig.\ref{fig:cd}, the pixel values of $I_{_{DAB}}$ can not describe the biomarker stain intensity. We propose to use a scheme described in Eq.\ref{eq:lalabel} to assign stain intensity values to pixels: 

\begin{eqnarray}
\label{eq:lalabel}
\nonumber{\lefteqn{ I_{la}(m,n)= }} \\
&\begin{cases}
I_l(m,n) & \condition[]{if $I_{_{DAB}}(m,n)$ is positive} \\
255 + (255 - I_l(m,n)) & \condition[]{if $I_{_{DAB}}(m,n)$ is negative}

\end{cases}
\end{eqnarray}
where $I_{la}(m,n)$ is the stain intensity description image.

The idea is that for the positive stain pixels, $I_{la}(m,n)$ is the same as the luminance component of the original image in order to preserve the morphology of the positive nuclei; for the negative stain pixels, $I_{la}(m,n)$ will have a higher value for strongly stained pixels (darker blue colour) and a lower value for weakly stained pixels (lighter blue colour). In order to separate the positive and negative pixel values clearly, we add an offset of 255 to the negatively stained pixels (most negative stain pixels will have a high $I_l(m,n)$ and positive stain pixels will have a low $I_l(m,n)$, the value of positive and negative pixels will be clearly separated in $I_{la}(m,n)$). Therefore, the larger $I_{la}(m,n)$ is, the weaker is the stain, the smaller $I_{la}(m,n)$ is, the stronger is the stain. When $I_{la}(m,n)$ is below or equal to 255, it is a positive stain pixel. In this way, we have obtained an image which gives a continuous description of the stain intensity of the image. Instead of setting artificial boundaries to separate the different degrees of stain intensity, we have now a continuous description of the stain intensity (see Fig.\ref{fig:color_thres}). Note that the pixel values of final image are normalized to the range from 0 to 1.

\subsection{Nuclei and Tumour Maps}
\label{sec:ntm}
As discussed above, the important information pathologists use to come up with the H-Score is the number of nuclei and the number of tumour nuclei in the TMA image. We therefore need to extract these two pieces of information and we use two separate FCNs, one for segmenting all nucleus and the other for segmenting tumour nucleus only.

To segment the tumour region, we use our own manually pixel-wise labelled tumour TMA images to train the FCN. While for segmenting general nuclei which detects both tumour and non-tumour nuclei, we utilize a transfer learning strategy to train another FCN. For general nuclei detection, the training data is obtained from three different datasets: immunofluorescence (IIF) stained HEp-2 cell dataset \cite{hobson2016hep}, Warwick hematoxylin and eosin (H\&E) stained colon cancer dataset \cite{sirinukunwattana2016locality}, and our own DAB-H TMA images. Since these three image sets are stained with different types of biomarker, we transform the colour image into grayscale for training. Training on a mixed image set could help to reduce overfitting on limited medical dataset and further boost the performance and robustness \cite{chen2016dcan}.

For both the general nuclei detection network and the tumour nuclei detection network, we use the symmetric U shape network architecture (U-Net) \cite{ronneberger2015u} with skip connection. The high resolution features from the contracting path are combined with the output from the upsampling path, which allows the network to learn the high resolution contextual information. The loss function is designed according the Dice coefficient as:
\begin{equation}
L_{mask} = - log \frac{2 \sum_{m,n} \omega \tau}{\sum_{m,n} \omega^2 + \sum_{m,n} \tau^2},
\end{equation}
where $\omega$ is the predicted pixel and $\tau$ is the ground truth. 

\subsection{The H-Score Prediction Framework}
The overview of the H-Score prediction framework is illustrated in Fig.\ref{fig:overall_flow}. It consists of three stages: 1) Nuclei segmentation, tumour segmentation, and stain intensity description; 2) Constructing the Stain Intensity Nuclei Image (SINI) and the Stain Intensity Tumour Image (SITI); 3) Predicting the final histochemical score (H-Score) by the Region Attention Multi-column Convolutional Neural Network (RAM-CNN). The rationale of this architecture is as follows: as only the number of nuclei, the number of tumour nuclei and the stain intensity of the tumour nuclei are the useful information for predicting H-Score, we therefore first extract these information. Rather than setting artificial boundaries for the categories of stain intensity, we retain a continuous description of the stain intensity. Only the information useful for predicting the H-Score is presented to a deep CNN to estimate the H-Score of the input image. This is in contrast to many work in the literature where the whole image is thrown to the CNN regardless if a region is useful or not for the purpose.   
\begin{figure*}[t]
	\centering
	\includegraphics[width=12cm,height = 6.5cm]{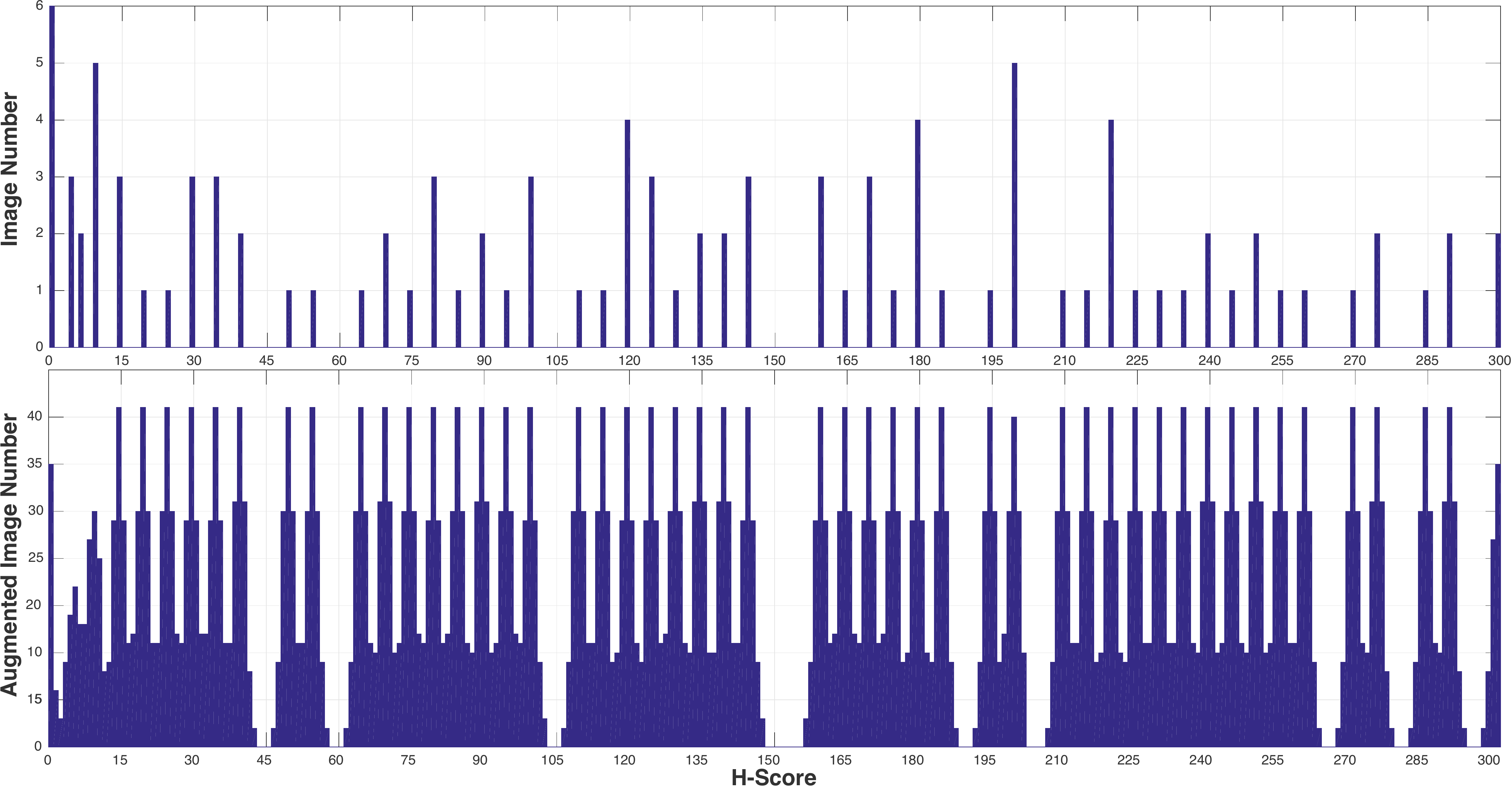}
	\caption{ The top graph is the original dataset label histogram; The bottom is the augmented label histogram.}
	\label{fig:data}
\end{figure*}

The detail of the first stage have been described in Section \ref{sec:sid} and \ref{sec:ntm}. As illustrated in Fig.\ref{fig:overall_flow}, an input TMA image $I(m,n)$ is processed by the tumour detection network which will output a binary image mask, $T(m,n)$, marking all the tumour nuclei, where $T(m,n) = 1$ if $I(m,n)$ is a part of a tumour nuclei and $T(m,n) = 0$ otherwise; by the general nuclei detection network which will output another binary image mask, $G(m,n)$, marking all tumour and non-tumour nuclei, where $G(m,n) = 1$ if $I(m,n)$ is a part of a nuclei and $G(m,n) = 0$ otherwise; and by the colour deconvolution and stain intensity labelling operation of Equation (8) to produce the stain intensity description image $I_{la}(m,n)$. 
In the second stage, we construct SINI and SITI by multiplying the nuclei mask image $G(m,n)$ and tumour mask image $T(m,n)$ with the stain intensity description image $I_{la}(m,n)$, i.e. $SINI = I_{la}(m,n) \times G(m,n)$, and  $SITI = I_{la}(m,n) \times T(m,n)$. Hence, all background pixels are zero, while only region of interests (ROI) are retained in SINI and SITI. All necessary information is preserved for histochemical assessment. Removing the background and only retaining ROI will enable the RAM-CNN convolutional layers to focus on foreground objects \cite{li2017not} which will significantly reduce computational costs and improve performance.
\begin{table}[!th]
	\begin{center}
		\begin{tabular}{|c||c|c|}
			\hline
			Layer &  \multicolumn{2}{c|}{Input / Filter Dimensions}\\
			\hline \hline
			Input & $512 \times 512\times1$ & $512 \times 512\times1$ \\
			\hline
			Conv1 & $8 \times 7 \times 7 $ & $8 \times 7 \times 7 $ \\
			MaxPooling & $ 2\times 2$ & $ 2\times 2$ \\
			\hline
			Conv2 & $16 \times 5 \times 5 $ & $16 \times 5 \times 5 $ \\
			MaxPooling & $ 2\times 2$ & $ 2\times 2$ \\
			\hline
			Conv3 &   \multicolumn{2}{c|}{$64 \times 3 \times 3 $ }\\
			MaxPooling & \multicolumn{2}{c|}{$ 2\times 2$}  \\
			\hline
			Conv4 &   \multicolumn{2}{c|}{$64 \times 3 \times 3 $ }\\
			MaxPooling & \multicolumn{2}{c|}{$ 2\times 2$}  \\
			\hline
			FC1& \multicolumn{2}{c|}{2048}\\
			\hline
			FC2 & \multicolumn{2}{c|}{1024}\\
			\hline
		\end{tabular}
	\end{center}
	\caption{The architecture of Region Attention Multi-channel Convolutional Neural Network (RAM-CNN).}
	\label{tab:racnn}
\end{table}

The proposed RAM-CNN is a deep regression model with dual input channels. The architecture of RAM-CNN is shown in Table \ref{tab:racnn}. Two inputs correspond to SINI and SITI respectively, and the input size is $512\times 512$. The parameters of the two individual branches are updated independently for extracting cell and tumour features respectively, without interfering with each other. The two pipelines are merged into one after two convolutional layers for H-Score prediction. The loss function for H-Score prediction is defined as:
\begin{equation}
L_{score} = \frac{1}{N} \sum_{i=1}^{N} \Vert F_{_{RAM}}(SINI_i,SITI_i) - l_i\Vert_2,
\end{equation}
where $F_{_{RAM}}(SINI_i,SITI_i)$ is the estimated score generated by RAM-CNN. $l_i$ is the ground truth H-Score.


\section{Experiments and Results}
\subsection{dataset}
The H-Score dataset used in our experiment contains 105 TMA images of breast adenocarcinomas from the NPI+ set \cite{rakha2014nottingham}. Each image contains one whole TMA core. The tissues are cropped from a sample of one patient which are stained with three different nuclei activity biomarkers: ER, p53, and PgR. The original images are captured at a high resolution of $40\times$ optical magnification, and then resized to 1024$\times$1024 pixels. The dataset is manually marked by two experienced pathologists with H-Score based on common practice. For each TMA core, the pathologists give the percentage of nuclei of different stain intensity levels, and then calculate the H-Score using Eq.\ref{eq:hscore}. The final label (H-Score) is determined by averaging two pathologists' scores, if the difference between two pathologists is smaller than 20. The dataset is available from the authors on request.

For training the general nuclei detection network, we transform Warwick H\&E colon adenocarcinoma \cite{sirinukunwattana2016locality} and NPI+ images to grayscale; the green channel was extracted from HEp-2 cell dataset \cite{hobson2016hep}. As HEp-2 cell images are IIF stained, the gray value should be inversed. 

\subsection{Data and Label Augmentation}
As in typical medical imaging applications, the dataset sizes are relatively small. In developing deep learning based solutions, it is a common practice to augment the training dataset for training. The training images for general nuclei detection network and tumour detection network are augmented by randomly cropping sub-images as input samples. For the H-Score dataset, rotation with random angles and randomly shifting the image horizontally and vertically within 5\% of image height and width are performed to augment the training set. 

As shown in the top row of Fig.\ref{fig:data}, the distribution of the label (H-Score) in the original dataset is unbalanced, some labels (H-Scores) have far more samples than others. Furthermore, one of the biggest problems is that because we have only limited number of samples, the H-Score values are discrete and discontinuous. There are many gaps between two H-Scores that has no data. Also, the values of the TMA image score given by the pathologists have a quantitative step-size of 5. Therefore, if an image has a score of 125, it means it has a value of around 125, the values in the vicinity of 125, i.e., 126 or 124 should also be suitable for labelling that image. In order to solve the ambiguity issue, we introduce Distributed Label Augmentation (DLA) which was inspired by the work of \cite{gao2017deep,zhou2012multi}. 
\begin{figure*}[!th]
	\centering
	\begin{minipage}[th]{1.2\linewidth}
		\includegraphics[height=4.4cm]{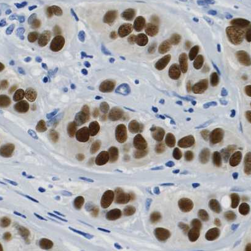}
		\includegraphics[height=4.4cm]{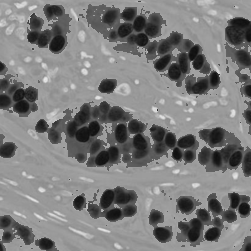}	
		\includegraphics[height=4.4cm]{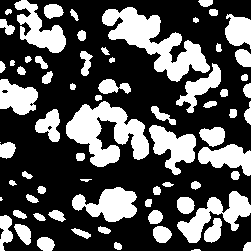}
		\includegraphics[height=4.4cm]{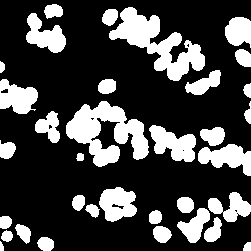}
	\end{minipage}
	\hspace{0.5\textwidth} 
	\\	
	\begin{minipage}[th]{1.2\linewidth}
		\includegraphics[height=4.4cm]{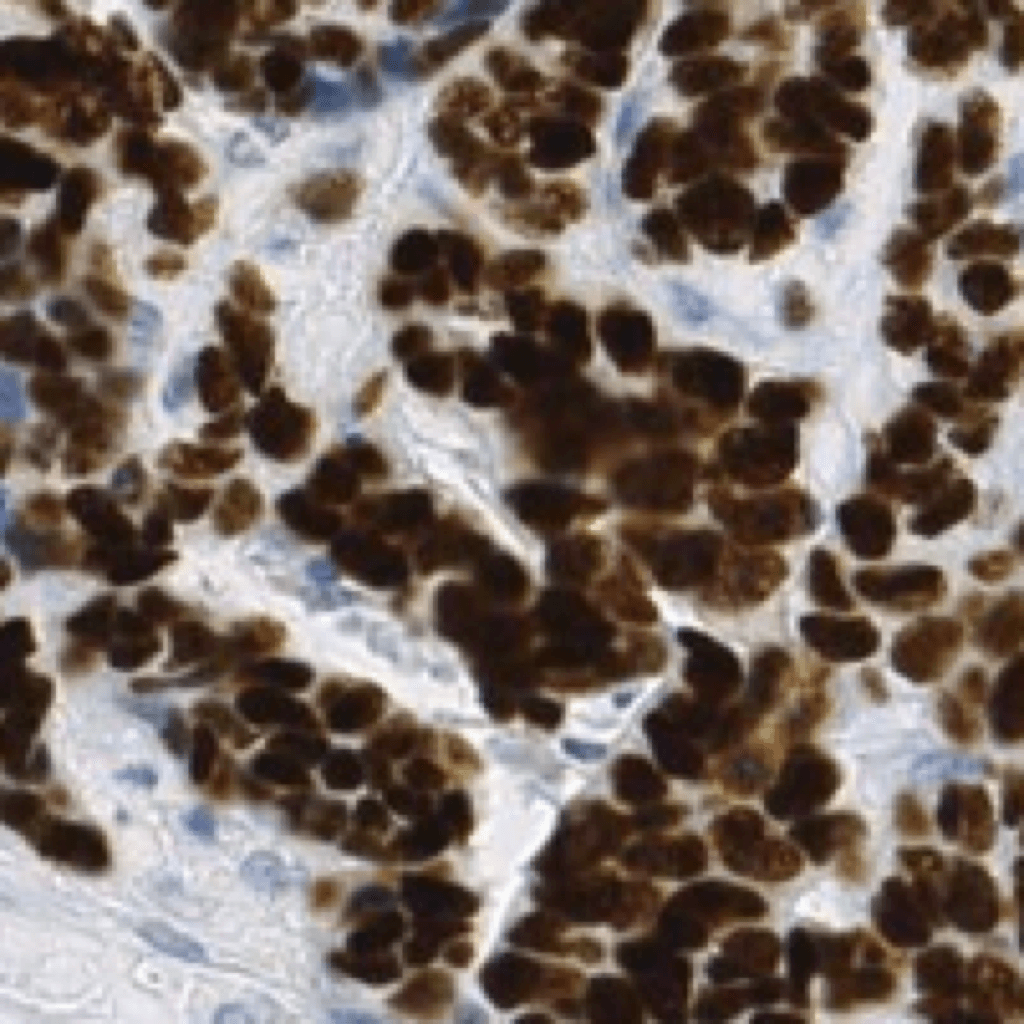}
		\includegraphics[height=4.4cm]{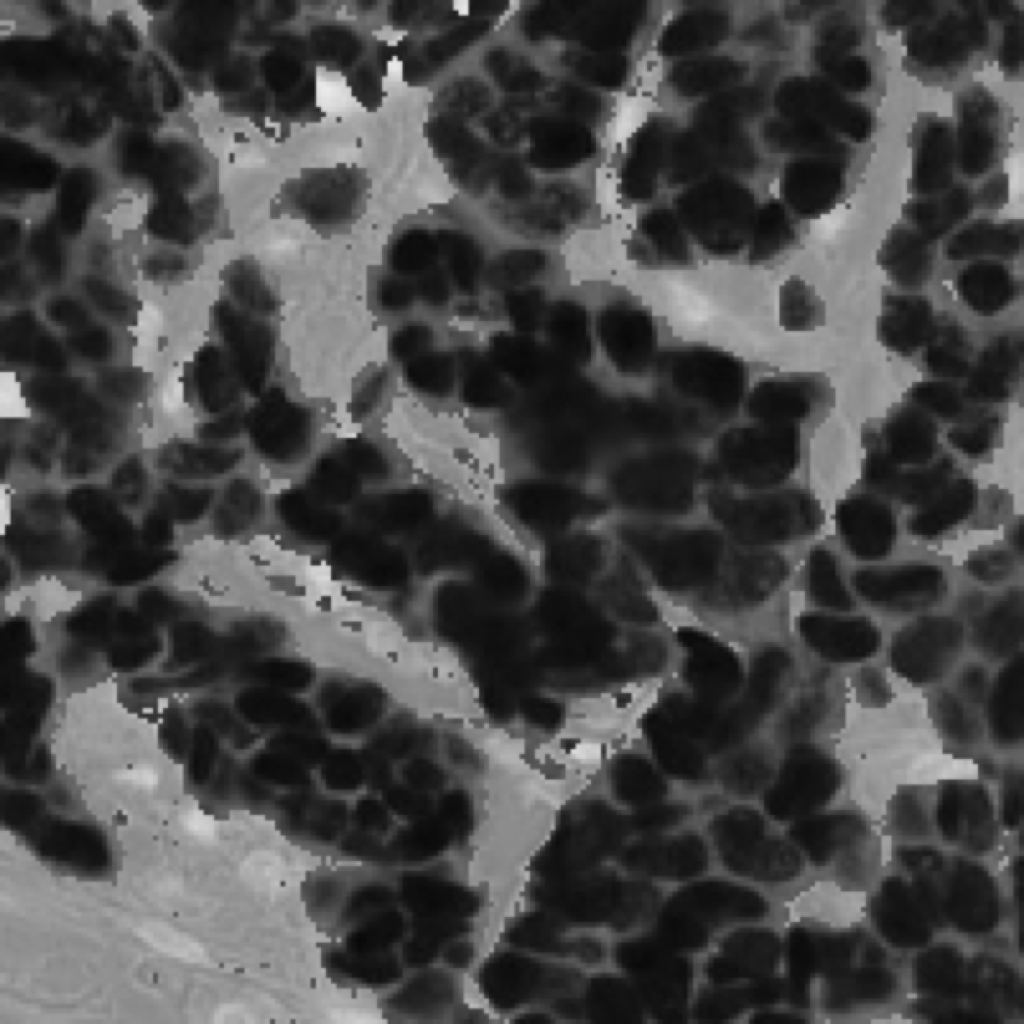}	
		\includegraphics[height=4.4cm]{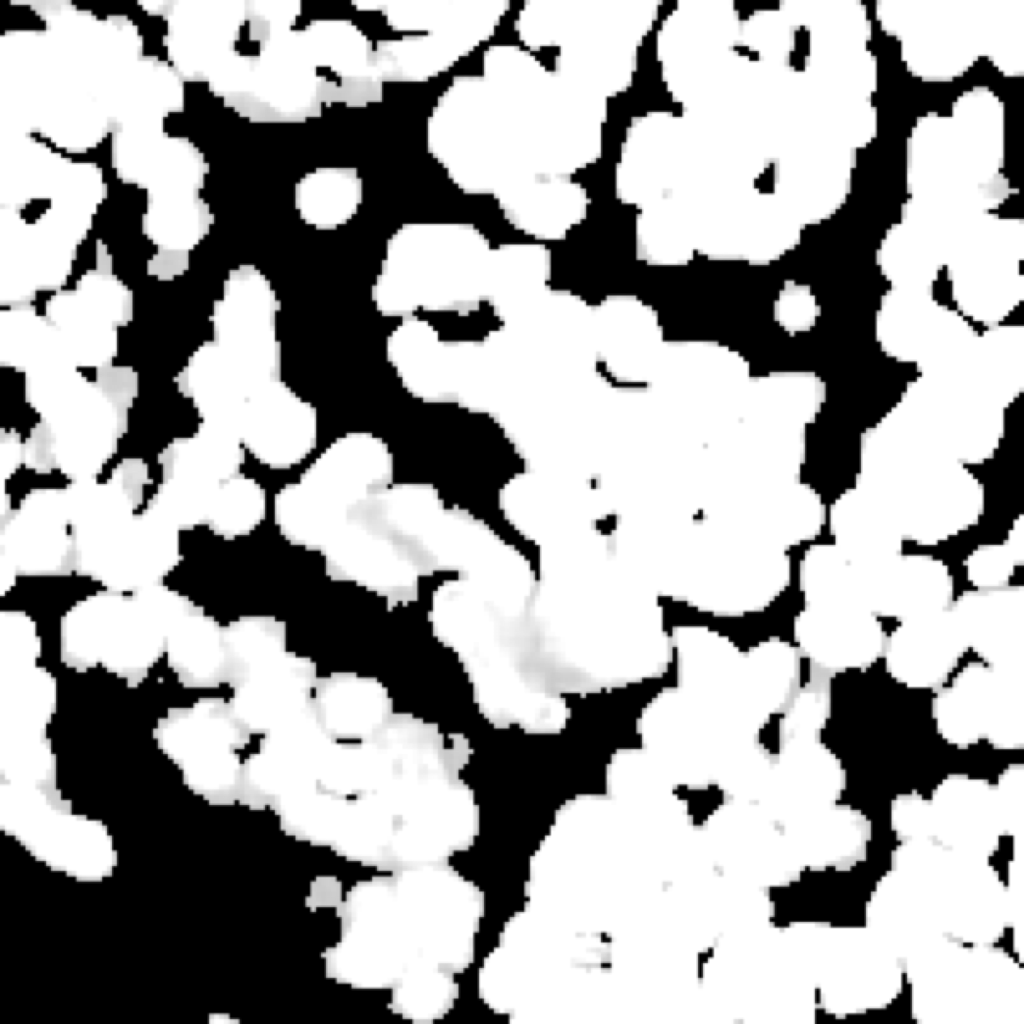}
		\includegraphics[height=4.4cm]{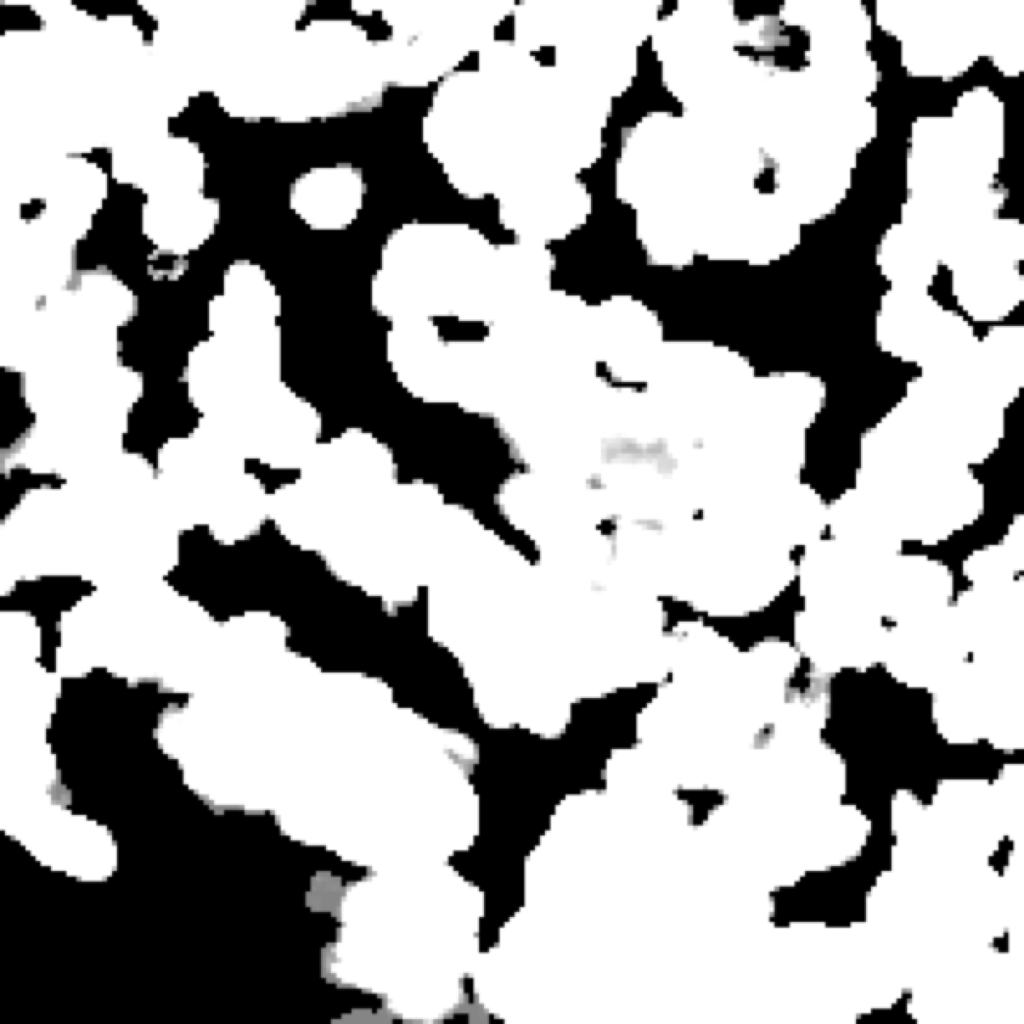}
	\end{minipage}
	\caption{Examples of intermediate images in the automatics H-Score prediction pipeline. From left to right: the original RGB image, luminance labelled stain intensity image, nuclei mask image, and tumour mask image respectively.}
	\label{fig:mid_result}
\end{figure*}

In the traditional regression method, a given dataset $\{(I_1,l_1), (I_2,l_2),\cdots,(I_D,l_D)\}$ pairs the instance $I_d$ for $1 \leq d\leq D$ with one single $l_d$ from the finite class label space $ \mathcal{L} = \left\{ l^0, l^1,\cdots, l^C \right\}$, where $C$ is the label size (\emph{e.g.}, $C = 301$ for H-Score). In this paper, the label is augmented so that one instance is associated with a number of labels. Formally, the dataset can be described as $\{(I_1,Y_1), (I_2,Y_2), \cdots, (I_D,Y_D)\}$, and $Y_d \subseteq \mathcal{Y}$ is a set of labels $\{y^{(1)}_d, y^{(2)}_d, \cdots, y^{(S)}_d \}$, where $S$ is the augmented label number for $I_d$. $y_d^{(s)}$ is sampled repeatedly from $\mathcal{L}$ based on a probability density function of following Gaussian distribution:
\begin{equation}
\label{eq:gau}
p(y^{(s)}_d = l^c) = \frac{1}{\sigma \sqrt{2\pi}} exp( - \frac{(l^c -\mu)^2}{2\sigma^2})
\end{equation}
where $\mu$ is the mean which equal to $l_d$ and $\sigma$ is standard deviation. Thus, $\sum_{s=1}^{S} p(y^{(s)}_d) = 1$ for each original TMA image. Consequentially, for an image $x_i$ from the augmented training set, its ground truth labels are assigned by repeatedly sampling from $\mathcal{L}$ according to Eq.\ref{eq:gau}. The augmented label histogram is shown at the bottom row of Fig.\ref{fig:data}.

\subsection{Implementation Details}
The network architecture for both the tumour nuclei detection and general nuclei detection models is the same as the U-Net \cite{ronneberger2015u} with a input size of $224\times224$. The filter size of tumour detection net is half narrower than that of general cell detection net. All networks use rectified linear unit (ReLU) activation function for the convolutional layer. The final cell and tumour region maps are predicted using sliding window.

A leave 5 out cross validation strategy is used for RAM-CNN model training, which means that in each round of testing, we randomly sample 5 TMAs as testing and the other 100 TMAs as training images. As explained previously, the training set is augmented via rotation and shift. The images are firstly resized to 512$\times$512 before fed into the RAM-CNN. We set $\sigma = 0.9$ to generate the H-Score distribution for ground truth label augmentation. We also add dropout layers after two fully connected layers with the rates of 0.3 and 0.5 respectively. The regression network is optimized by Adam \cite{kingma2014adam} with an initial learning rate of 0.001.

\begin{figure*}[!th]
	\centering
	\subfloat[NAP]{%
		\includegraphics[width=4.4cm,height = 4cm]{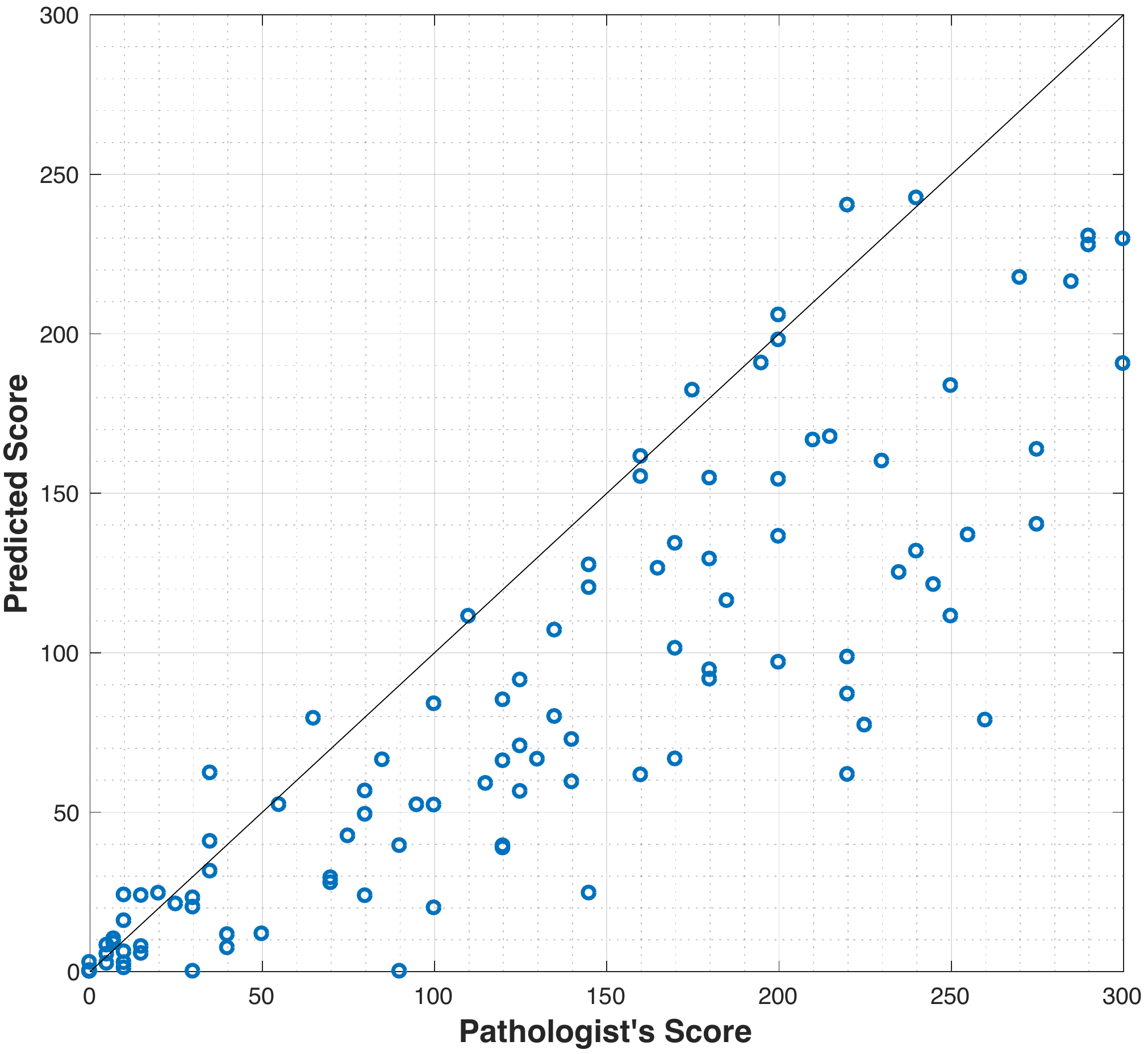}}\
	\subfloat[NNP]{%
		\includegraphics[width=4.4cm,height = 4cm]{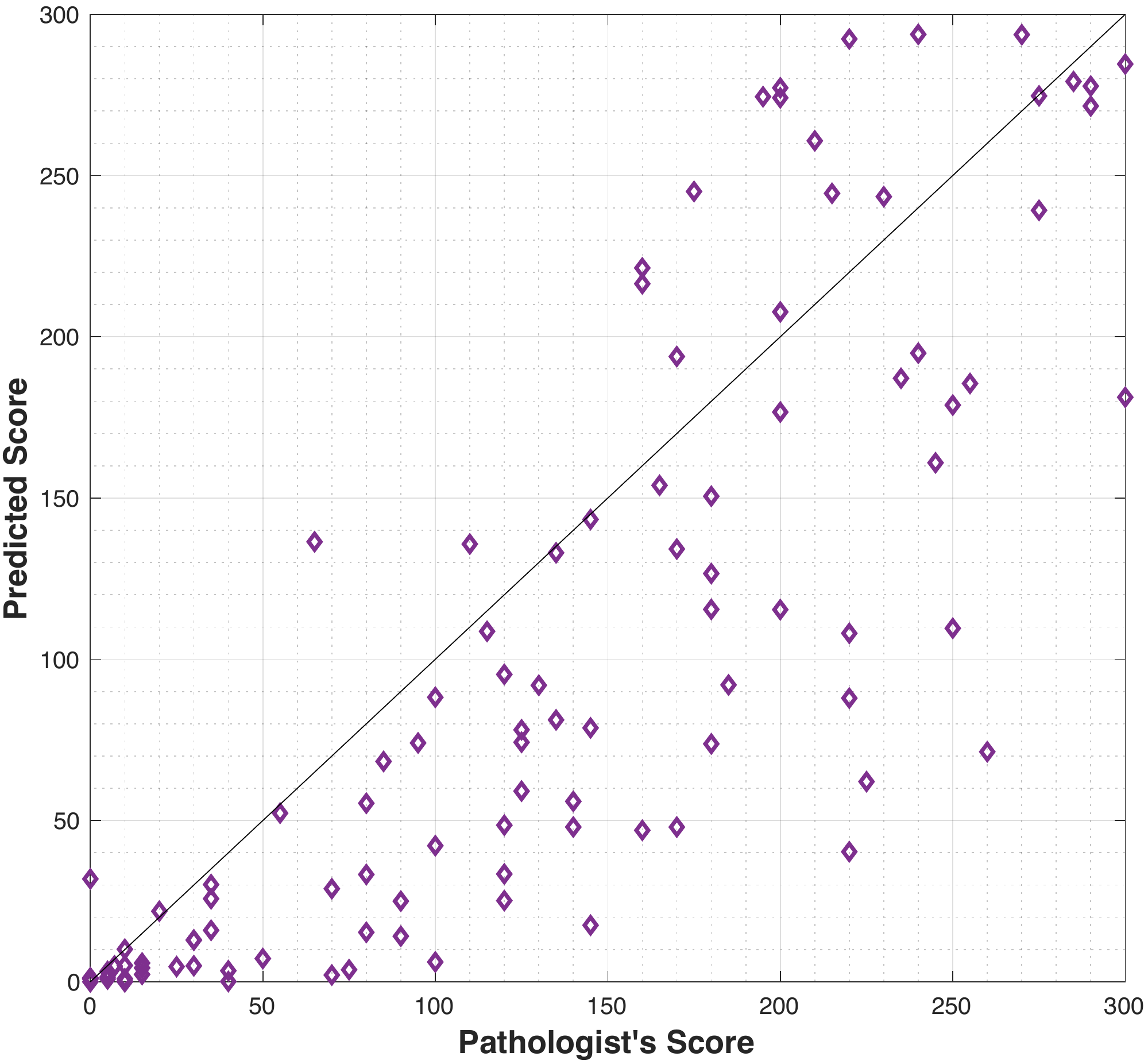}}\\
	\subfloat[RGB-CNN]{%
		\includegraphics[width=4.4cm,height = 4cm]{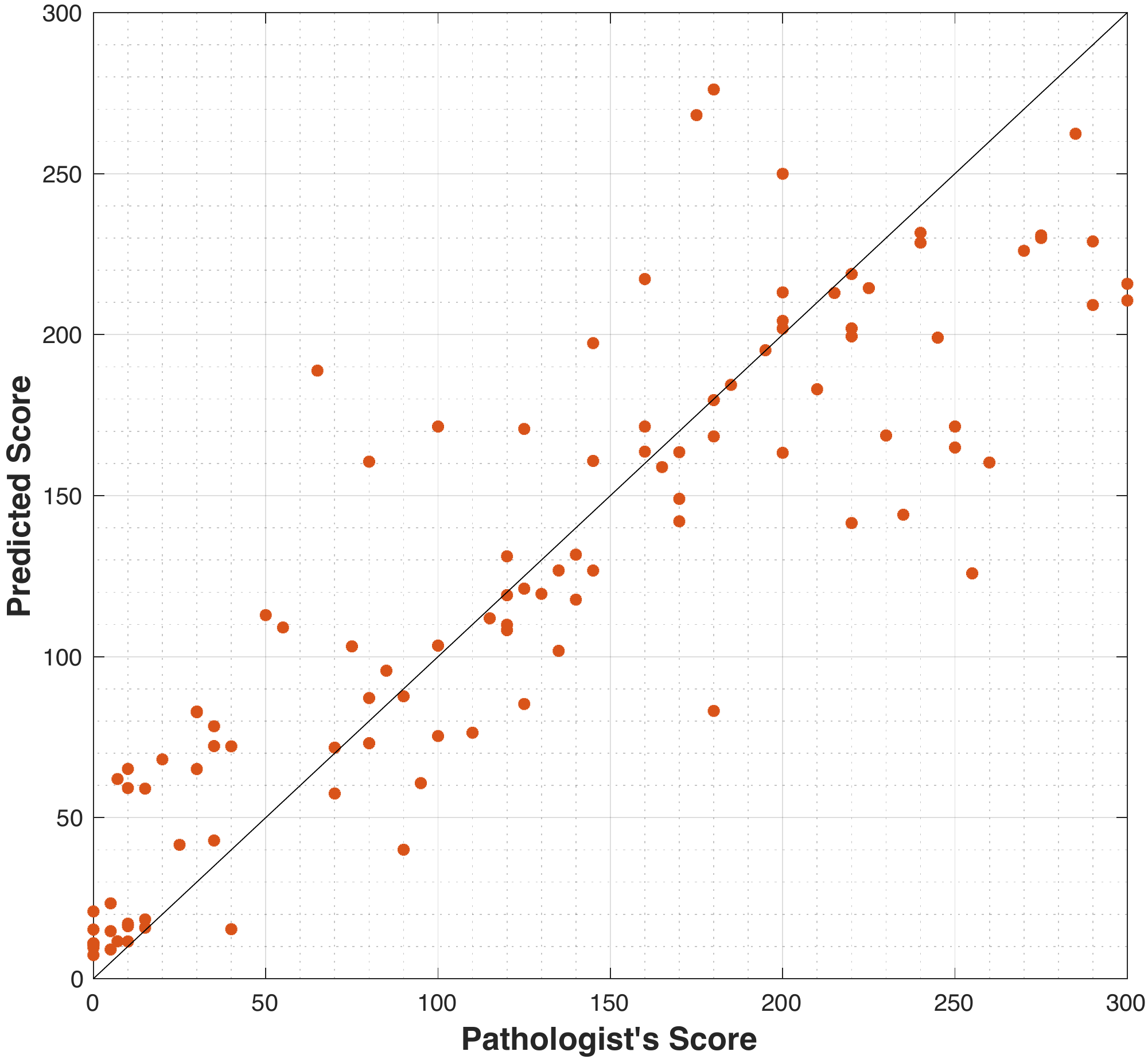}}\
	\subfloat[RA-CNN]{%
		\includegraphics[width=4.4cm,height = 4cm]{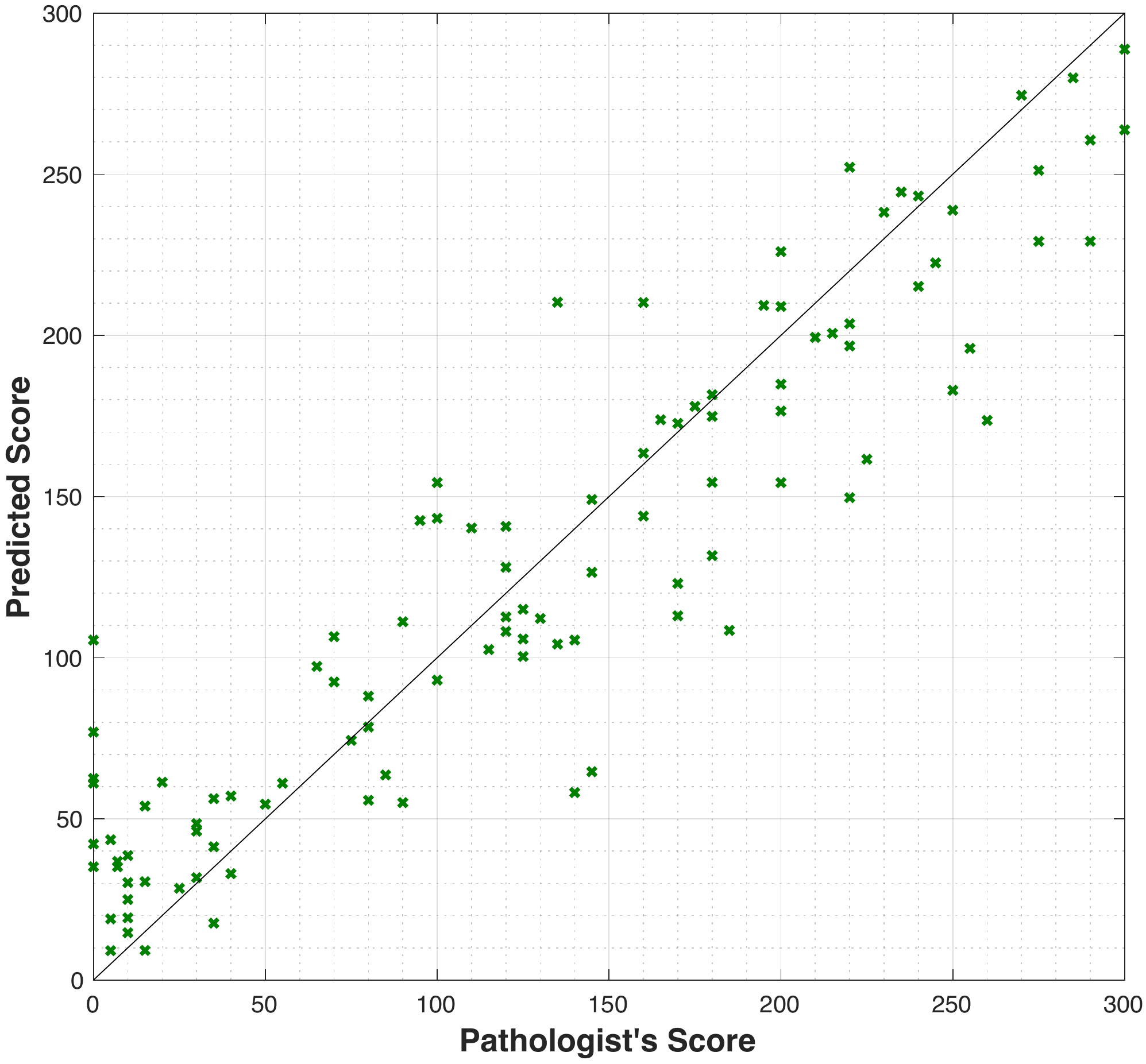}}\
	\subfloat[RAM-CNN]{%
		\includegraphics[width=4.4cm,height = 4cm]{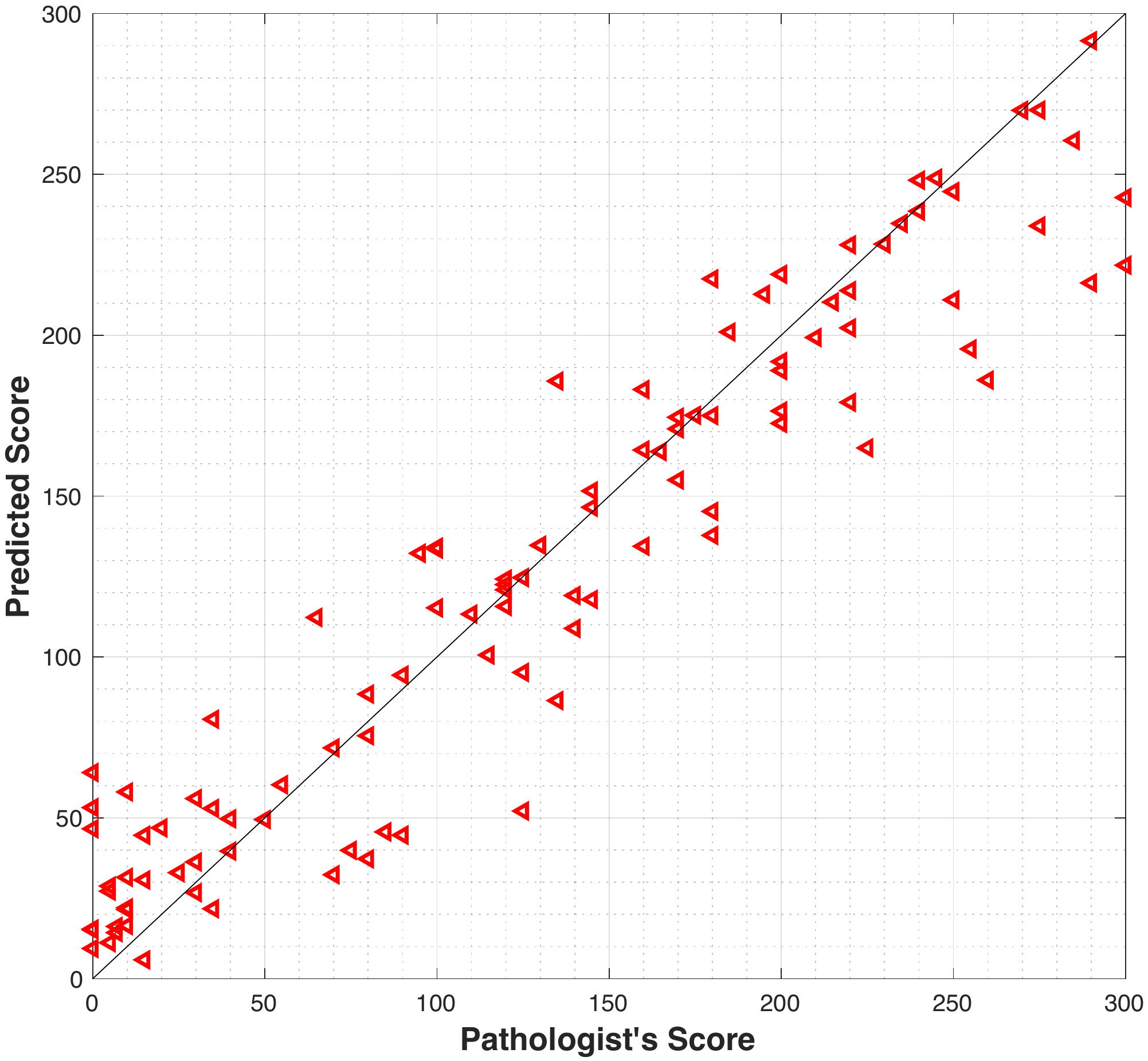}}
	
	\caption{Scatter plots of the predicted scores of different models vs the pathologists' manual scores. }
	\label{fig:scat_method}
\end{figure*}

\subsection{Results  and Discussions}
\subsubsection{Experimental Results}
\label{sec:exp_res}
Fig.\ref{fig:mid_result} shows some examples of the intermediate images in the automatic H-Score prediction pipeline. It is seen that the luminance labelled stain intensity image marks a sharp distinction between positive and negative stains. This shows that our maximum a posteriori (MAP) classifier based Luminance Adaptive Multi-Thresholding (LAMT) method \cite{liu2016luminance} can reliably separate positive DAB stains for a variety of images. It also shows that our stain intensity labelling strategy can preserve the morphology of the nuclei, separate positive and negative stains while retaining a continuous description of the positive stain intensity.

\begin{figure}[!th]
	\centering
	\subfloat[]{%
		\includegraphics[height = 3.9cm,width=4.2cm]{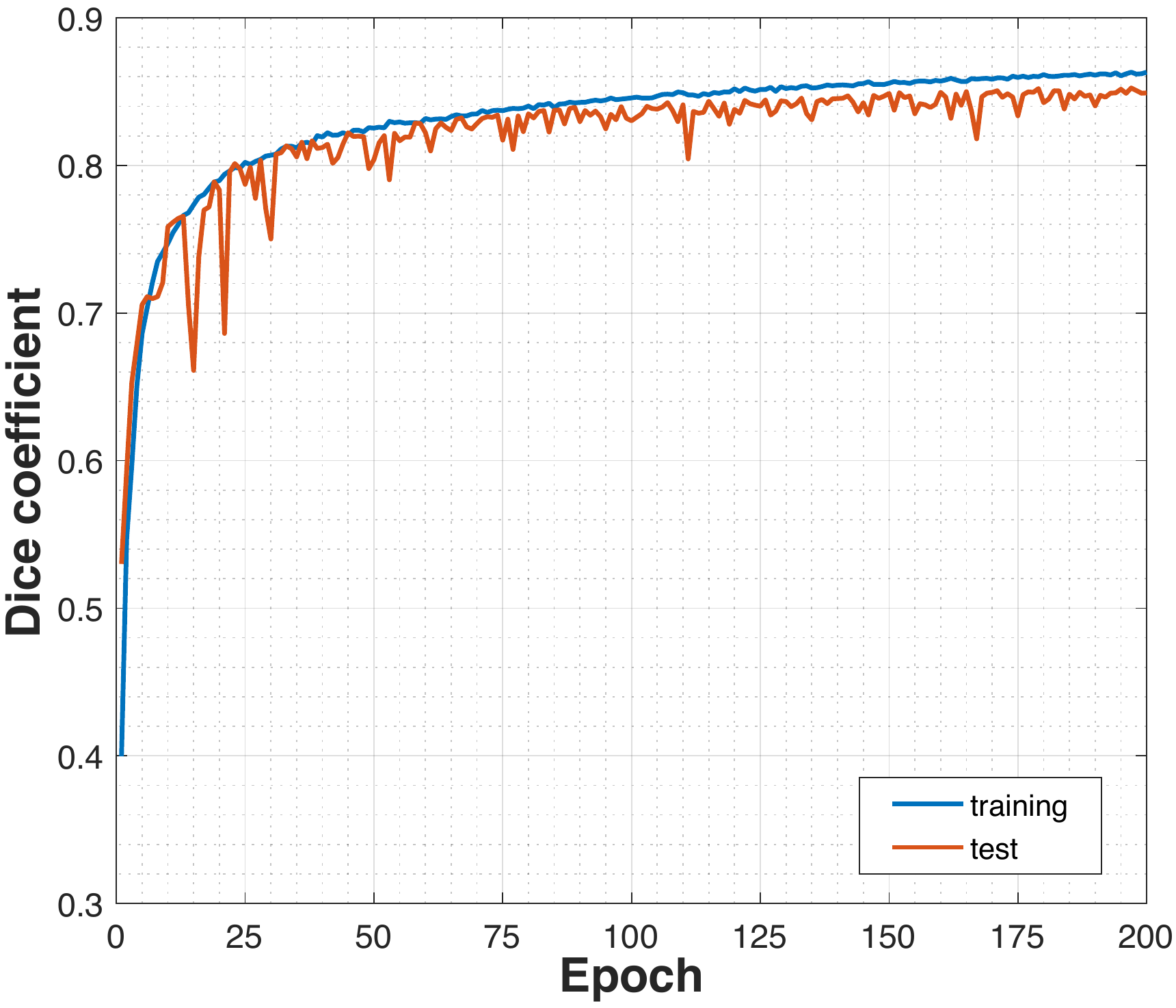}}		
	\subfloat[]{%
		\includegraphics[height = 3.9cm,width=4.2cm]{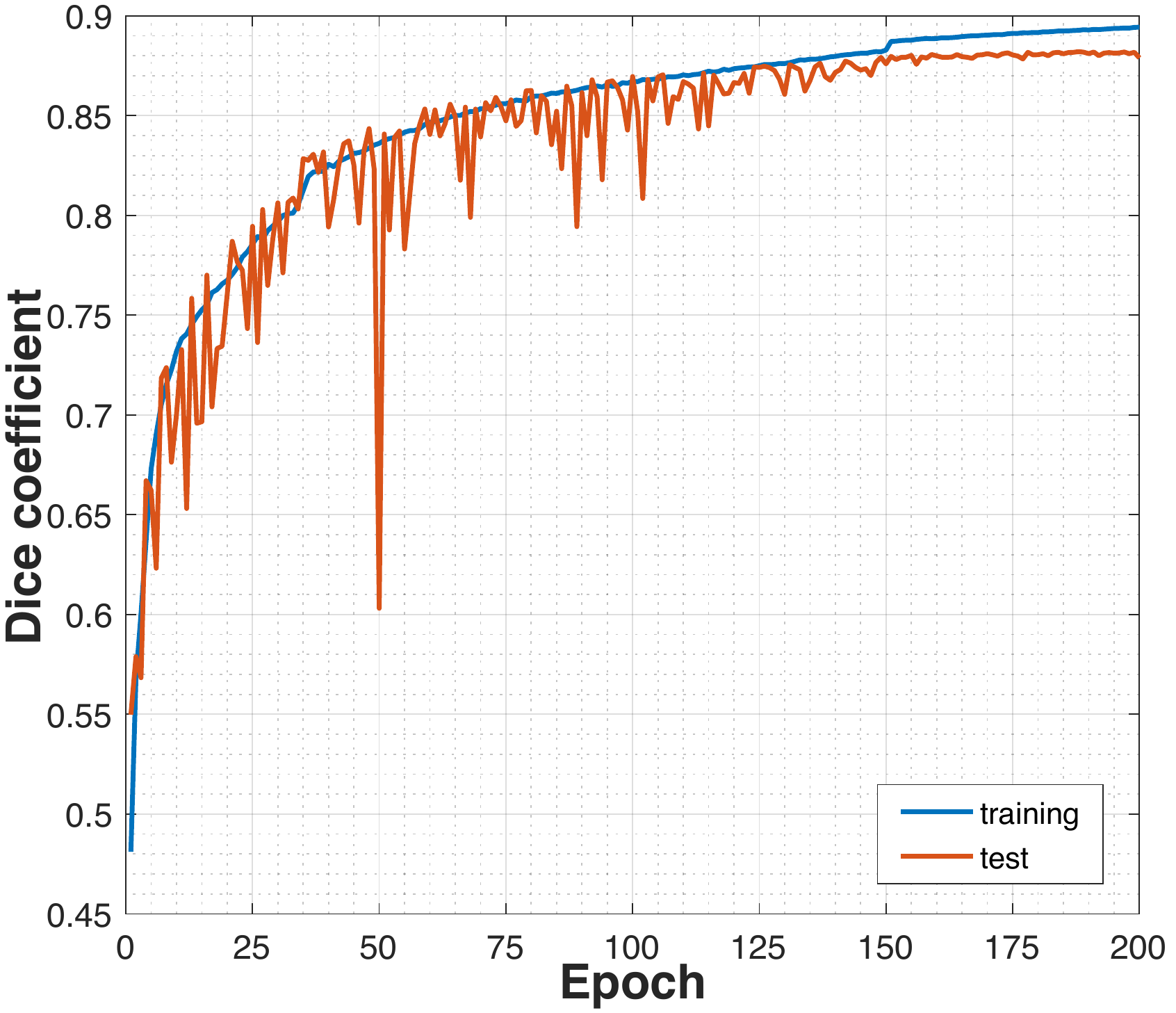}}
	
	\caption{Training results for general nuclei detection network (a) and tumour nuclei detection network (b). }
	\label{fig:detect_result}
\end{figure}

Fig.\ref{fig:detect_result} shows the training curves of the Dice coefficient for the general nuclei detection network and the tumour nuclei detection network respectively. Both networks converged after 170 epochs respectively. The nuclei mask images (see Fig.\ref{fig:mid_result}) show that the deep convolutional network trained with mixed datasets using transfer learning can successfully detect the nuclei in our H-Score dataset. The tumour segmentation network is able to identify tumour region from normal tissues. It is worth noting that the ground truth masks for the two detection networks are different. All nucleus in Warwick colon cancer images are labelled with circular masks with a uniform size, while tumour region masks are pixel level labelled. Therefore, the final predicted maps generated by the two networks for the same nucleus are different. In addition, it is found that the mask dilation become evident with the increase of DAB stain intensity. One possible reason is that the strong homogeneous stain makes the nuclei texture and edge feature difficult to extract. 

To evaluate the performance of our proposed RAM-CNN and the two H-Score relevant images SINI and SITI, we compare our model with two traditional single input pipeline CNNs: RGB-CNN and RA-CNN (i.e., region attention CNN). The RGB-CNN takes the original RGB TMA image with the shape of $512\times512\times3$ as input, and output the H-Score prediction. To investigate the effect of multi-column architecture, we combine SINI and SITI as a two channel image of $512\times512\times2$ for the input of RA-CNN. The architectures of RGB-CNN and RA-CNN are the same as a single pipeline RAM-CNN (see Table.\ref{tab:racnn}).

We also calculate the H-Score using Eq.\ref{eq:hscore} based on the nuclei area percentage (NAP) and nuclei number percentage (NNP). Specifically, the luminance labelled stain intensity description image $I_{la}$ is first calculated according to the description in Section \ref{sec:sid}. The pre-defined thresholds \cite{trahearn2016simultaneous} are utilized for categorizing the pixels into different DAB-H stain intensity groups. For the NAP method, the predicted H-Score is calculated according to the percentages of area in different stain intensity groups. NNP employs the NIH ImageJ tool \cite{schneider2012nih} with a multi-stage cell segmentation technique \cite{shu2013segmenting} for cell detection. The detected cells are classified into \emph{unstained}, \emph{weak}, \emph{moderate}, and \emph{strong} groups using the pre-defined thresholds for H-Score calculation . 

\begin{table}[!th]
	\begin{center}
		\begin{tabular}{|c|c|c|c|c|}
			\hline
			Model & MAE & SD & CC & P value\\
			\hline\hline
			NAP & 47.09 & 46.03& 0.87& $< $ 0.001\\
			\hline
			NNP & 46.48 & 55.18& 0.82& $< $ 0.001\\
			\hline
			RGB-CNN & 32.01 & 44.46& 0.87& $< $ 0.001\\
			\hline
			RA-CNN & 27.22 & 35.72& 0.92& $< $ 0.001\\
			\hline
			RAM-CNN & \textbf{21.33}& \textbf{29.14}& \textbf{0.95} & $< $ 0.001\\
			\hline
			Human & 20.63 & 30.55 & 0.95 & $< $ 0.001\\
			\hline
		\end{tabular}
	\end{center}
	\caption{Performance comparison with different regression models. The last line (Human) are difference between the H-Scores given by the two pathologists.}
	\label{tab:mae}
\end{table}

In this paper, Mean Absolute Error (MAE), Standard Deviation (SD) and the correlation coefficient (CC) between the predicted H-Score and the average H-Score of the two pathologists are used as the evaluation metrics. As a reference, we also calculate the MAE, SD and CC between the H-Scores given by the two pathologists of all original diagnosis data. Results are shown in Table.\ref{tab:mae}. 
\begin{figure*}[!t]
	\centering
	\subfloat[0-49]{%
		\includegraphics[width = 3cm]{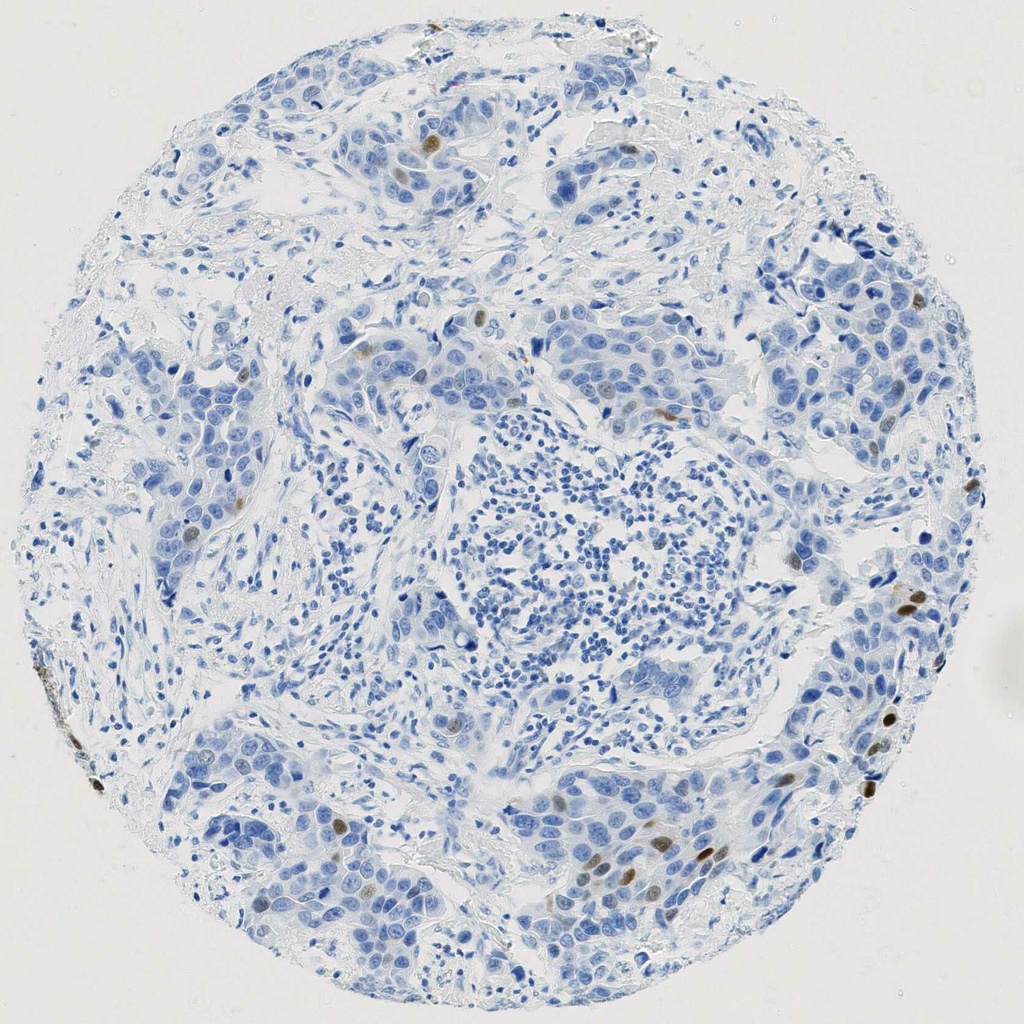}}
	\subfloat[50-99]{%
		\includegraphics[width = 3cm]{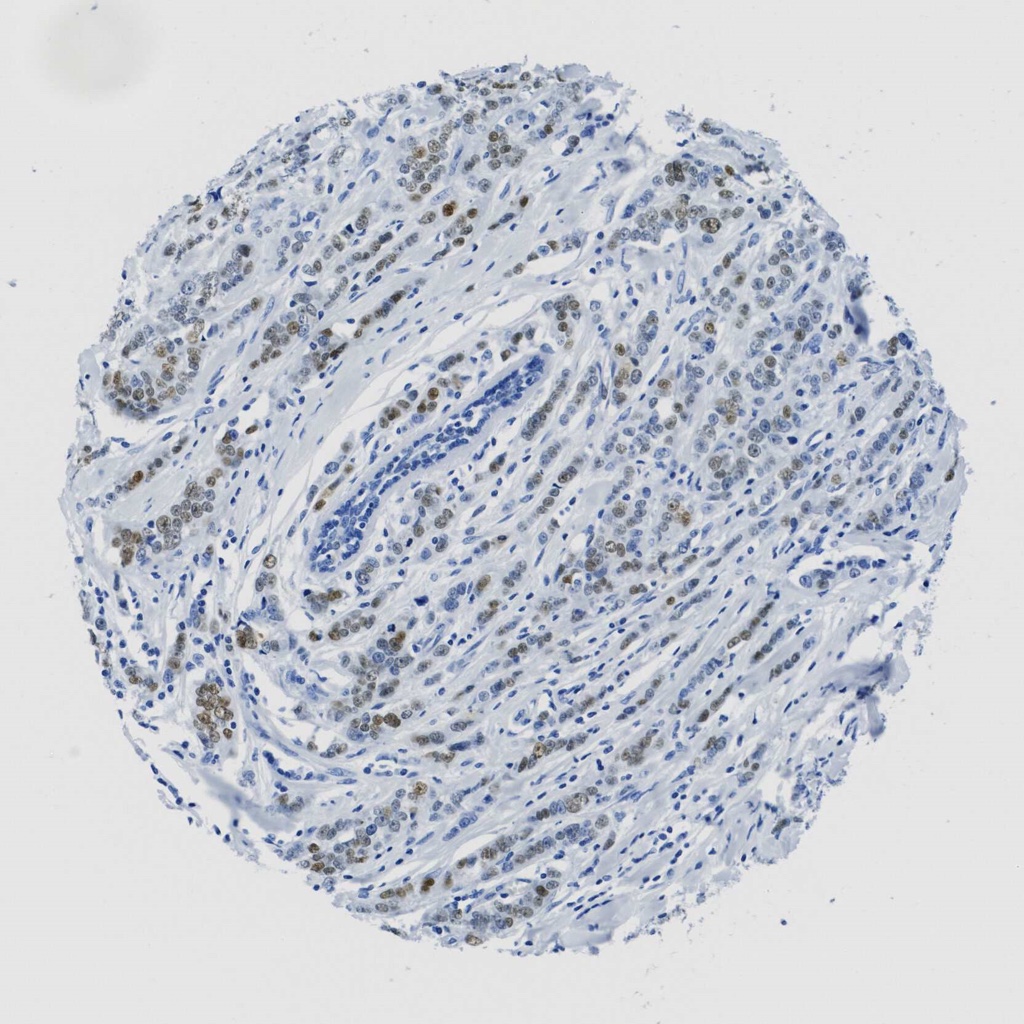}}
	\subfloat[100-149]{%
		\includegraphics[width = 3cm]{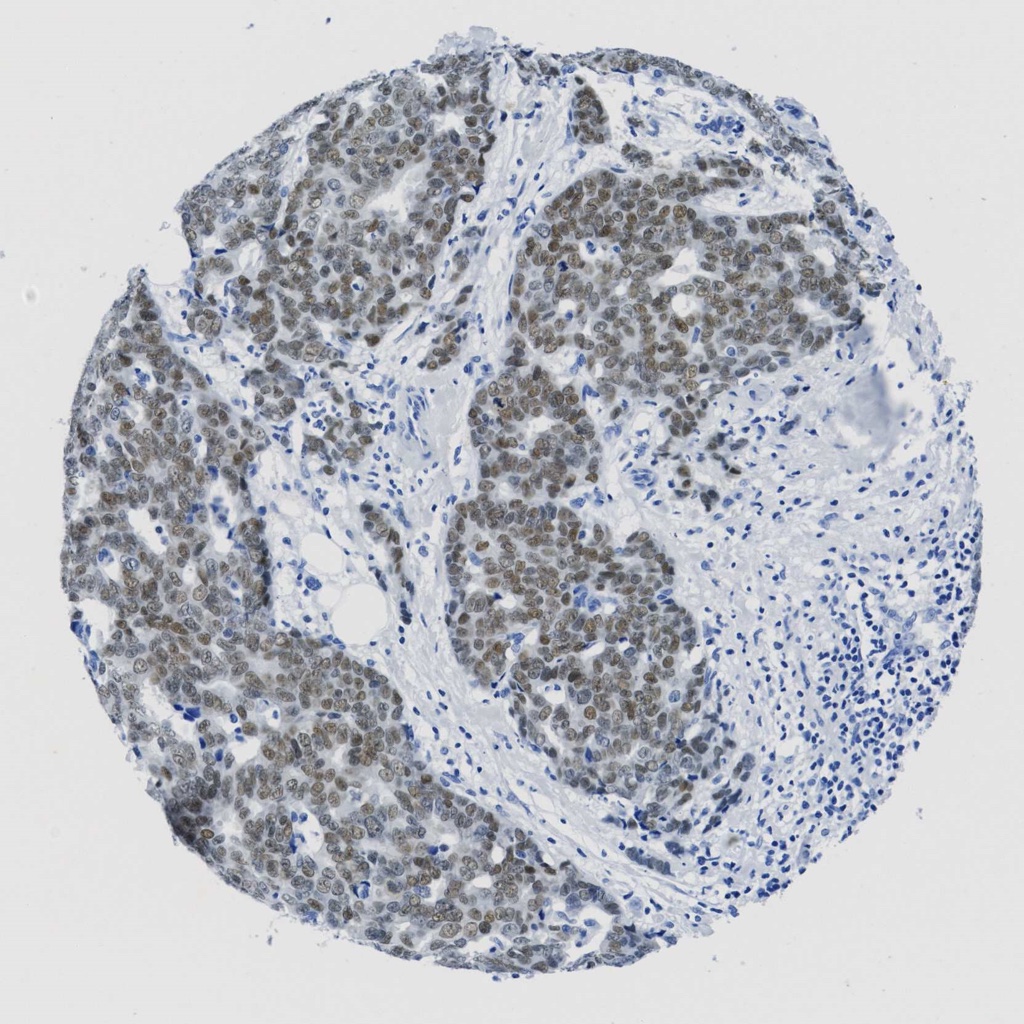}}
	\subfloat[150-199]{%
		\includegraphics[width = 3cm]{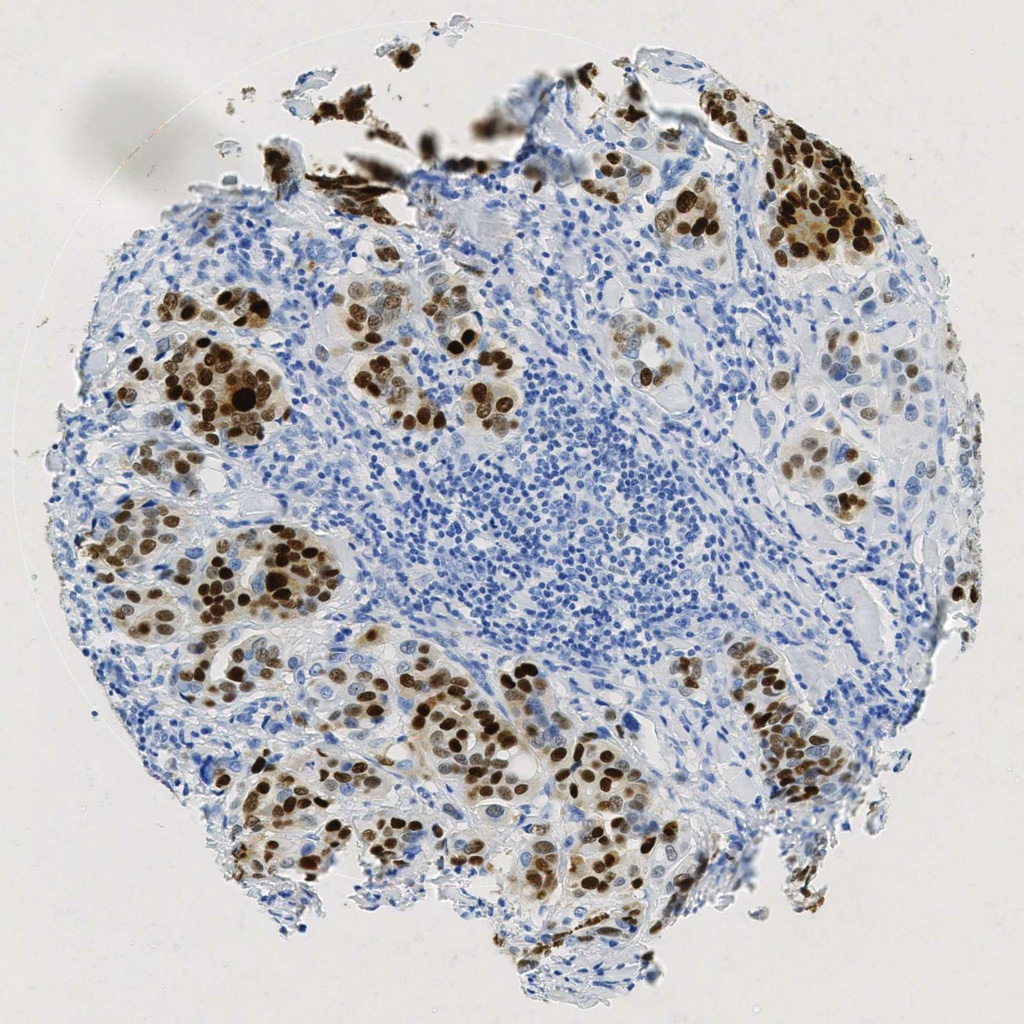}}
	\subfloat[200-249]{%
		\includegraphics[width = 3cm]{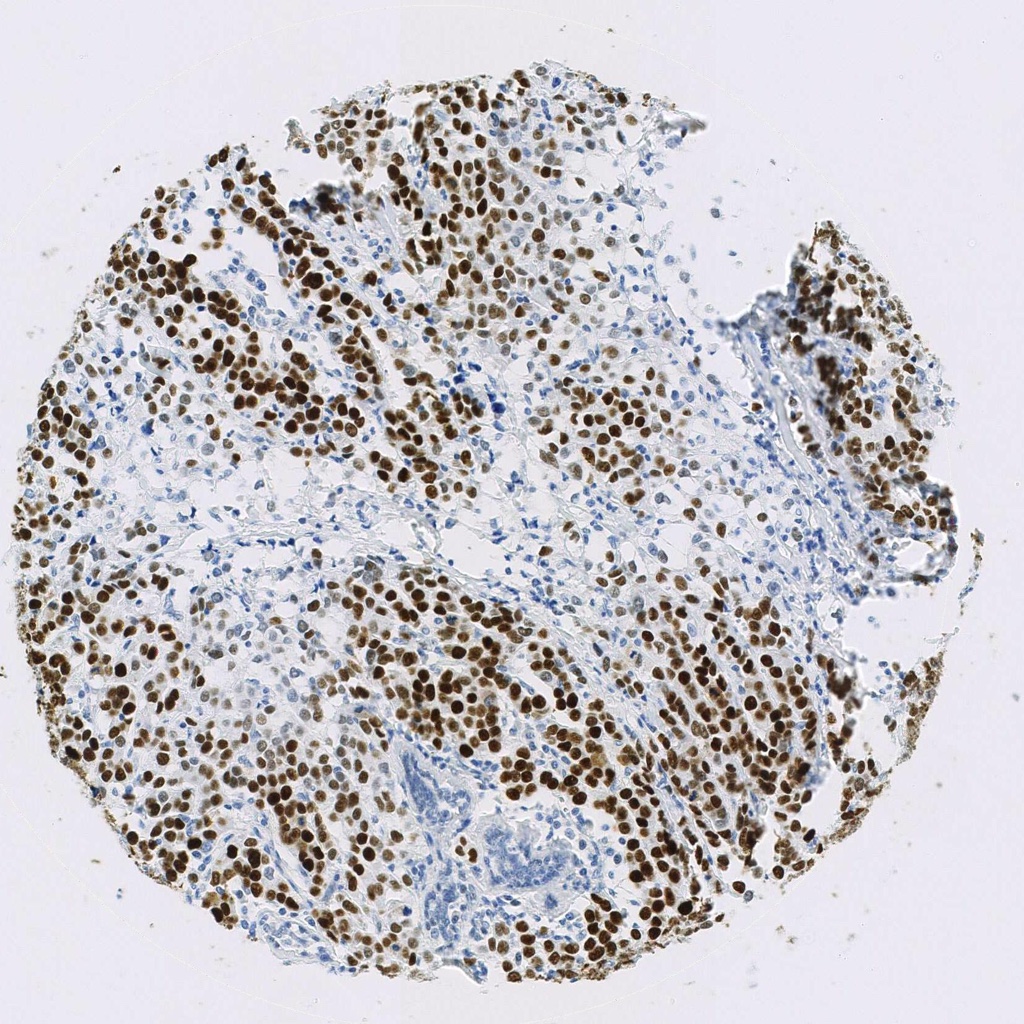}}
	\subfloat[250-300]{%
		\includegraphics[width = 3cm]{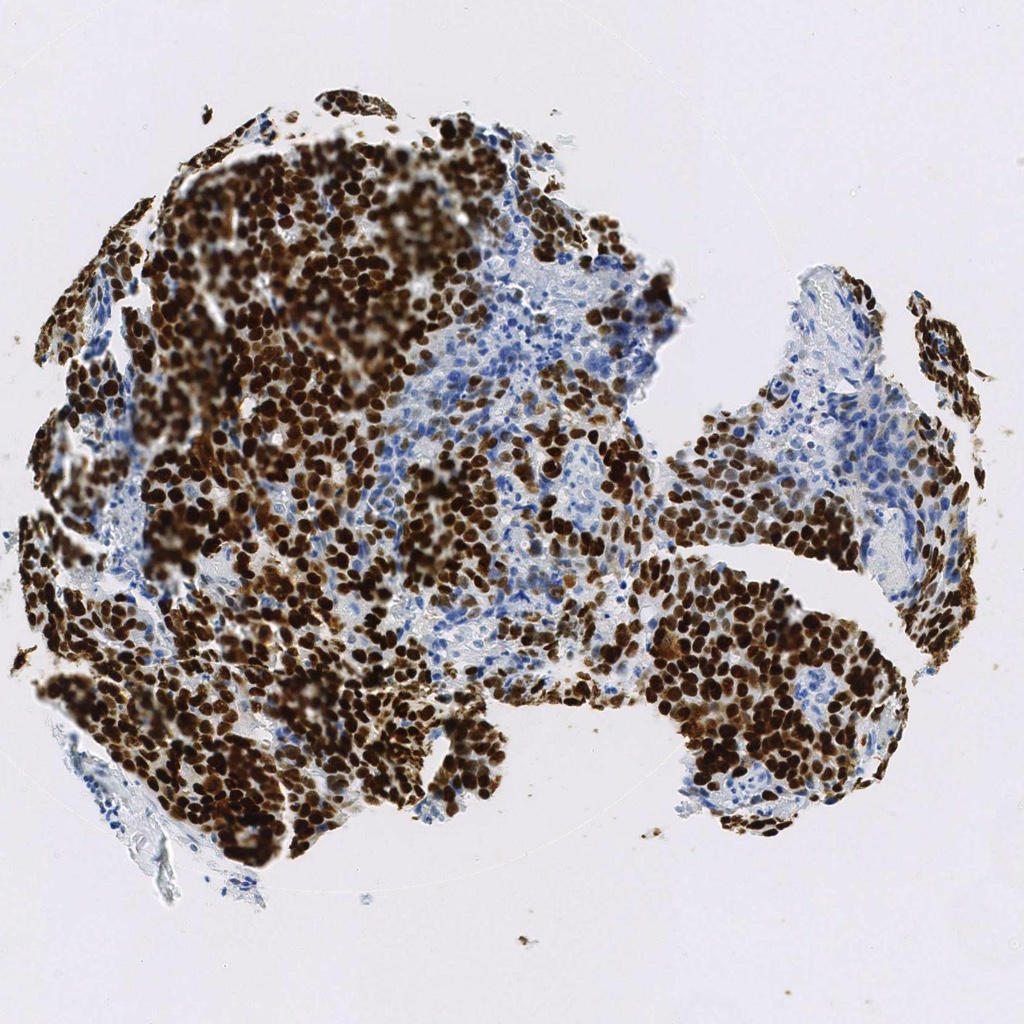}}
	\caption{Example TMA images extracted from different H-Score groups.}
	\label{fig:group_img}
\end{figure*}

As can be seen, the NAP based prediction gives the highest MAE with large deviations in the cross validation, which is followed by NNP. Our RAM-CNN framework achieves the lowest prediction error (21.33); a traditional CNN setting with the proposed SINI and SITI as input gives the second lowest prediction error (27.22). This verifies the effectiveness of our proposed approach to filtering out irrelevant pixels and only retain H-Score relevant information in SINI and SITI. All deep learning based methods outperform NAP and NNP by a large margin. To investigate the statistical significance of automatically predicted H-Scores, the correlation of the predicted and those of the pathologists scores and its P value are also calculated. The correlation between pathologists scores and those predicted by RAM-CNN is $0.95$ with a P value of $< 0.001$ which means there is strong evidence against the null hypothesis \cite{wasserstein2016asa}. 

It is interesting to observe that the difference between our RAM-CNN predicted H-Scores and the average of the two pathologists H-Scores (MAE = 21.33, SD = 29.14, CC = 0.95) are on par with the difference between the two pathologists (MAE = 20.63, SD = 30.55, CC = 0.95). While the MAE between the RAM-CNN and humans is slightly higher than that between humans, the SD between humans is higher than that between RAM-CNN and humans. The CC between humans and machine and that between humans are the same.  

Fig. \ref{fig:scat_method} illustrates the scatter plots between the model predicted scores and the pathologists' scores. Most of the predicted scores of NAP are lower than the ground truth. At the lower end, NNP predicted scores are lower than the ground truth while at the higher end the predicted scores are higher than the ground truth. 
These two methods are affected by several low-level processing components including the pre-defined stain intensity thresholds and the nuclei segmentation accuracy. Our proposed framework gives more accurate prediction results compared to traditional single pipeline CNN, further demonstrating that imitating the pathologists' H-Scoring process by only keeping useful information is an effective approach. 


\subsubsection{Discussions}

In this paper, we introduced an end-to-end system to predicted H-Score. To investigate the reason for scoring discrepancy between the proposed algorithm and the pathologists, we firstly
compare the H-Score prediction results for different biomarkers as shown in Table.\ref{tab:biomarkerCompare}. The proposed framework gives the best accuracy in all three biomarker images. The performances are slightly different for different biomarkers. This is to be expected because different markers will stain the tissues differently. 
Although the difference is not large, whether it will be useful to train a separate network for different biomarkers is something worth investigating in the future. 
\begin{table}[H]
	\centering
	\begin{tabular}{|c|c|c|c|}
		\hline
		Biomarker & ER    & P53   & PgR   \\
		\hline
		No. of TMA    & 32 &   33    &   40    \\
		\hline
		\hline
		NAP       & 42.02 & 50.68 & 48.17 \\
		\hline
		NNP       & 43.53 & 46.72 & 48.92 \\
		\hline
		RGB-CNN   & 24.90 & 31.57 & 38.19 \\
		\hline
		RA-CNN    & 25.43 & 23.39 & 31.82 \\
		\hline
		RAM-CNN   & 21.01 & 16.66 & 25.44\\
		\hline
	\end{tabular}
	\caption{Comparing MAE of different methods on three different biomarkers.}
	\label{tab:biomarkerCompare}
\end{table}
\begin{figure}[!th]
	\centering	
	\includegraphics[width = 8.5cm]{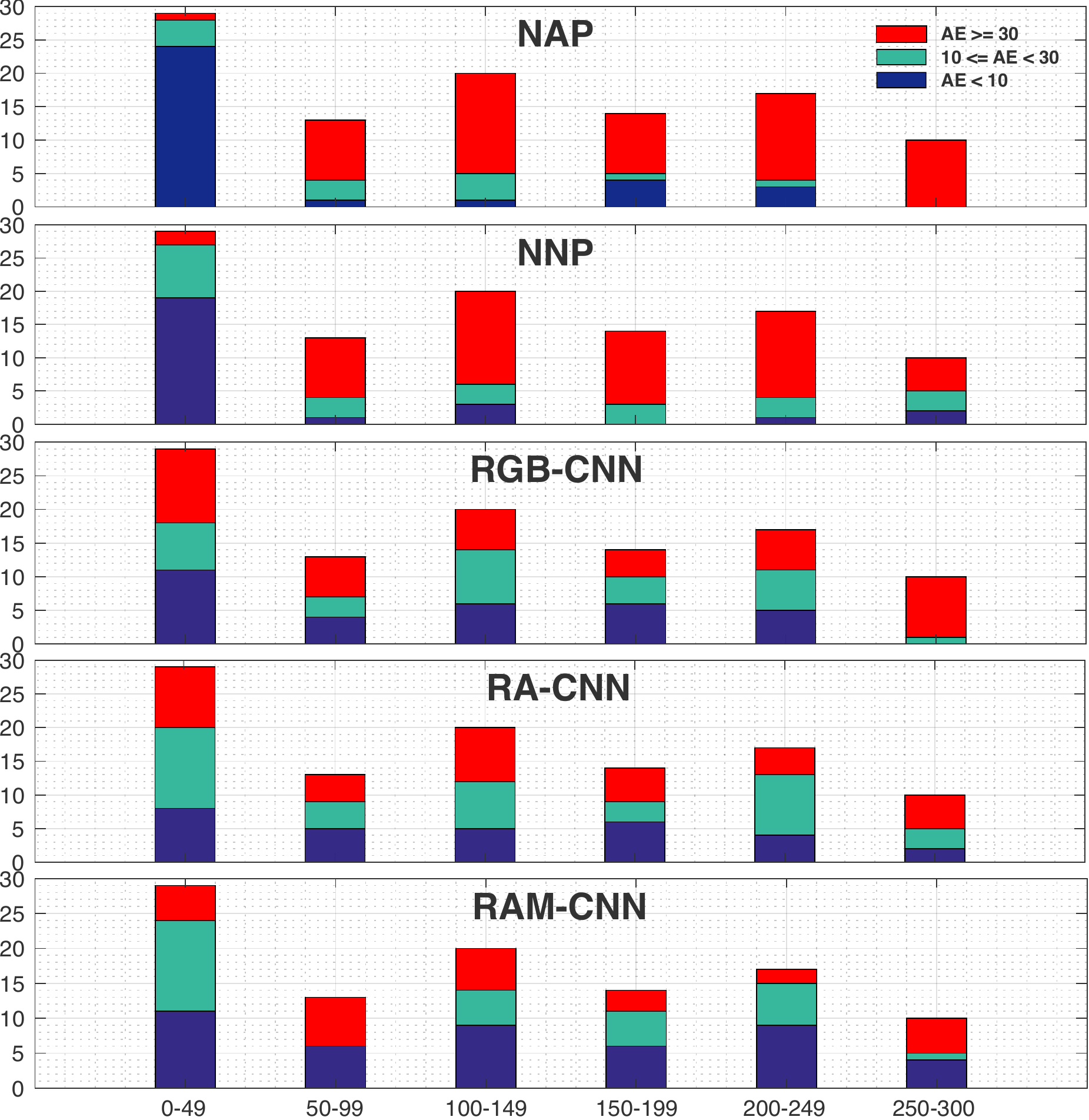}		
	\caption{Comparison of performances of different methods in different H-Score groups.}
	\label{fig:scoreComp}
\end{figure}

To see how the algorithms perform differently across the dataset, we divide the TMA images into 6 groups according to their pathologists' scores. Example TMA images of each group are illustrated in Fig.\ref{fig:group_img}. For each group, we count the number of TMAs with absolute error (AE) smaller than 10, between 10 and 30, and larger than 30 respectively. The results of different methods are shown in Fig.\ref{fig:scoreComp}. It is seen that in the low H-Score group of 0-49, traditional methods of NAP and NNP give more accurate predicted scores than CNN based methods. It is found that most low score TMAs are unstained or weakly stained as shown in Fig.\ref{fig:group_img}(a). The accurate predictions from NAP and NNP indicate that the predefined threshold for separating \emph{unstained} and \emph{weak} (see Fig. \ref{fig:color_thres}) is compatible with pathologists' criteria. The deep learning based methods do not set stain intensity thresholds explicitly and their performances across the six groups are relatively even. 

The accuracies of NAP and NNP decrease rapidly with the increase of the H-Score. As shown in Fig.\ref{fig:group_img}, the stain intensity and image complexity increase with the H-Score which directly affect the performance of traditional methods. The result also indicates that the pre-defined stain intensity thresholds for \emph{moderate} and \emph{strong} classes (see Fig. \ref{fig:color_thres}) are less compatible with the pathologists' criteria. Furthermore, the large coefficients of \emph{moderate} and \emph{strong} stain (see Eq.\ref{eq:hscore}) would magnify the errors of area and nuclei segmentation in NAP and NNP respectively.

Three deep learning based methods give worse results on the groups with fewer images (i.e., group 50-99 and 250-300), which indicates the importance of a large training data size. In addition, the uneven distribution of original dataset may also affect the predicted accuracy.

\begin{figure}[!th]
	\centering
	\subfloat[]{%
		\includegraphics[height = 3.8cm]{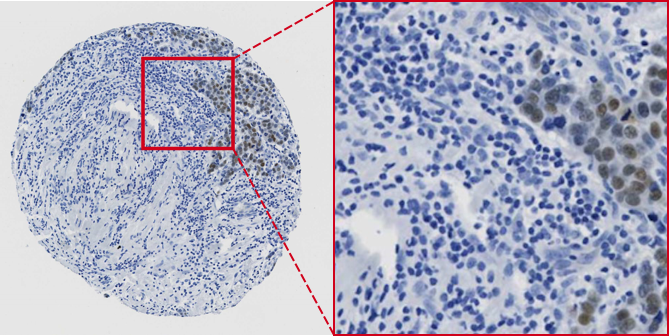}}\
	\subfloat[]{%
		\includegraphics[height = 3.8cm]{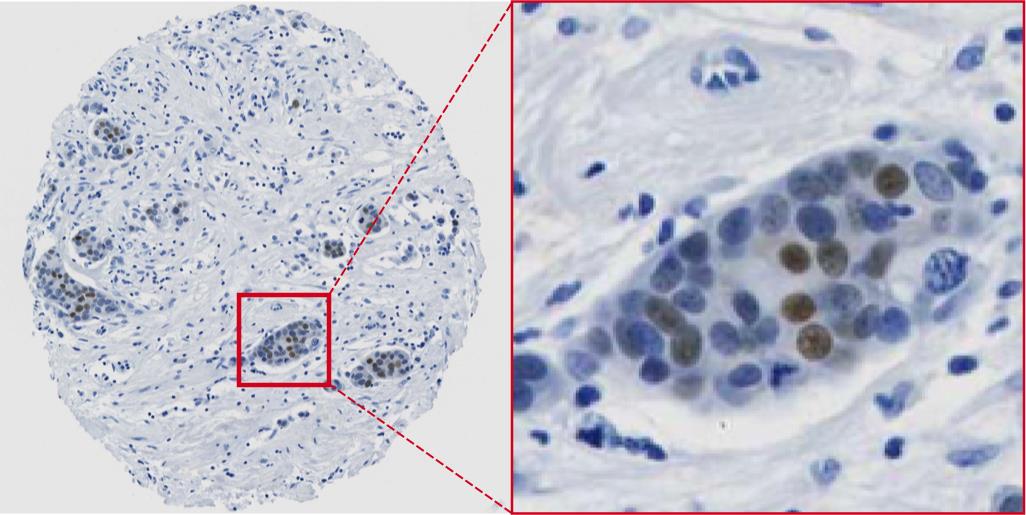}}
	\caption{Examples of accurately scored TMAs by proposed algorithm. The absolute errors generated by RAM-CNN of both (a) and (b) are smaller than 2.}
	\label{fig:pos_exa}
\end{figure}

\begin{figure*}[!th]
	\centering
	\subfloat[]{%
		\includegraphics[height = 3.8cm]{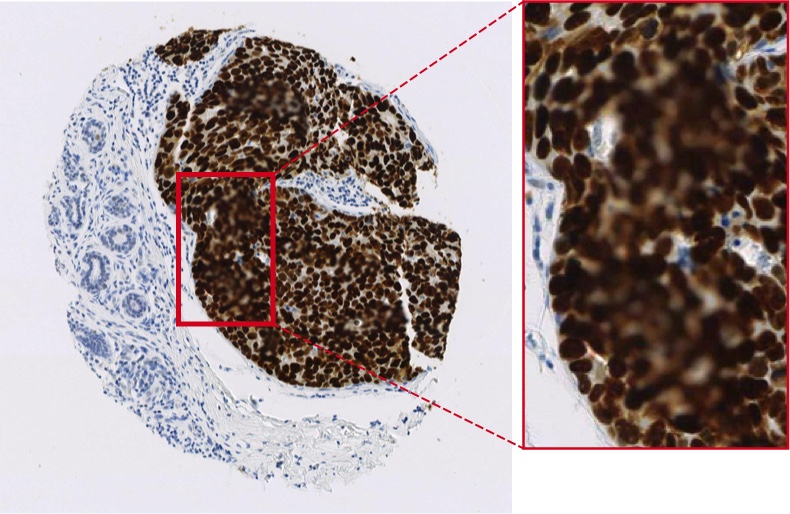}}\
	\subfloat[]{%
		\includegraphics[height = 3.8cm]{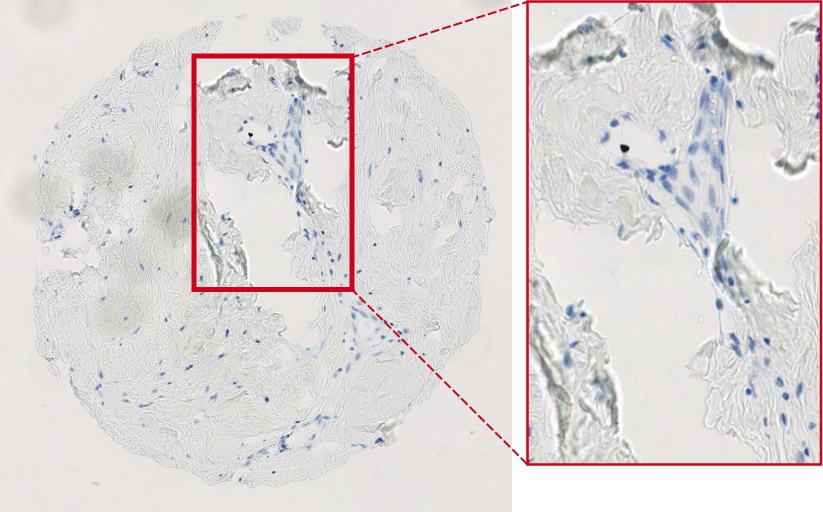}}\
	\subfloat[]{%
		\includegraphics[height = 3.8cm]{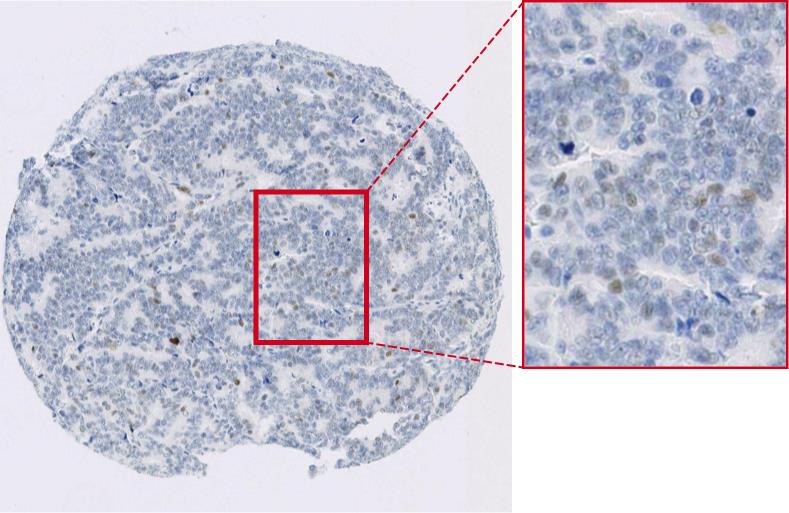}}
	\caption{Examples of sources of big scoring discrepancy between algorithm and pathologist. (a) out of focus; (b) tissue folds; (c) heterogeneity and overlapping.}
	\label{fig:err_exa}
\end{figure*}

We further analyse the TMAs individually to investigate the effect of image quality on the proposed algorithm. We found that for those TMAs where the tissues are clearly stained, and the cellular structure is clear without severe overlap (see Fig.\ref{fig:pos_exa}), our algorithm can give very accurate prediction. On the other hand, poor image quality causes errors. In the images that are most easily mis-scored by our algorithms, we found three significant characteristics as shown in Fig.\ref{fig:err_exa}. 

The TMA core in Fig.\ref{fig:err_exa}(a) contains large out-of-focus regions, which happens more commonly on strongly stained tissues. The blur regions directly affect the performance of nuclei segmentation, as well as the nuclei and tumour detection accuracy. They also hinder the final regression network from extracting topological and morphological information. 

Tissue folds (see Fig.\ref{fig:err_exa}(b)) occurs when a thin tissue slice folds on itself, and it can happen easily during slide preparation especially in TMA slides. Tissue-fold would cause out-of-focus during slide scanning. Furthermore, a tissue fold in a lightly stained image can be similar in appearance to a tumour region in a darkly stained image \cite{kothari2013eliminating}. Hence, the segmentation accuracy of colour deconvolution would be greatly affected in tissue-fold regions.

Heterogeneity and overlapping as shown in Fig.\ref{fig:err_exa}(c) also affect the automatic scoring performance. The stain heterogeneity gives rise to a large discrepancy of stain intensity in a single nucleus, and nuclei overlapping adds to the difficulty.

These three difficulties directly affect the predicted results of the proposed method, and we found that most large mis-scored TMAs contain one or more of these characteristics. We found that there were 9 low image quality TMAs in our dataset and if we exclude these 9 lowest-quality TMA images, the average MAE of our RAM-CNN is 18.79. Therefore, future works need to overcome these issues in order to achieve a high prediction performance. To solve the problem of out-of-focus, heterogeneity and overlapping, adding corresponding images in the training set to promote robustness is one potential quality assurance methods. In addition, the deep learning based scoring system can be developed to add nuclei number estimation function for accurate assessment. It is also necessary to add the function of automated detection and elimination of tissue-fold regions before H-Score assessment.

\section{Concluding Remarks}
In this paper, we have developed a deep learning framework for automatic end-to-end H-Score assessment for breast cancer TMAs. 
Experimental results show that automatic assessment for TMA H-Score is feasible. The H-Scores predicted by our model have a high correlation with H-Scores given by experienced pathologists. We show that the discrepancies between our deep learning model and the pathologits are on par with those between the pathologists. We have identified image out of focus, tissue fold and overlapping nuclei as the three major sources of error. We also found that the major discrepancies between pathologists and machine predictions occurred in images that will have a high H-Score value. These findings have suggested future research directions for improving accuracy. 



\bibliographystyle{IEEEtran}
\bibliography{mybib}
%

%




\end{document}